%% file: msc_report.tex
\documentclass[11pt,a4paper]{article}

\usepackage[english]{babel}
\usepackage[utf8]{inputenx}
\usepackage[T1]{fontenc}

\usepackage{nameref}
\usepackage{hyperref}
\hypersetup{colorlinks=false,pdfborder={0 0 0}}

\usepackage{graphicx}
\usepackage[pdf]{graphviz}
\usepackage{amsmath,amsfonts,amssymb,amsthm}

\usepackage{float}
\usepackage{framed}
\usepackage{mdframed}
\usepackage{tcolorbox}
\usepackage{longtable}
\usepackage{rotating}

\usepackage{setspace}
\usepackage{listings}

\theoremstyle{definition}
\newtheorem{exmp}{Example}[section]

\tcbset{colback=white,colframe=black}

\graphicspath{{figures/}{figures/r2rml/}{figures/npd/}{figures/aci/}}

\input{lstsettings}
\input{notation}
\providecommand{\keywords}[1]{\textbf{\textit{Keywords---}} #1}

\begin{document}

\title{\input{title}}
\author{Manuel Namici\vspace{0.5em}\\ DIAG, Sapienza, University of Rome\vspace{0.2em}\\ {\small\em manuel.namici@gmail.com}}
\date{}
\maketitle

\begin{abstract}
  \input{abstract}
\end{abstract}

\input{sec-intro}
\input{sec-r2rml}
\input{sec-npd-prep}
\input{sec-aci-prep}
\input{sec-conclusions}

\bibliographystyle{abbrv}
\bibliography{bibliography/bibliography}
\nocite{*}

\end{document}

%% file: lstsettings.tex
\lstdefinelanguage{SPARQL}%
{morekeywords={ACTION,AS,BETWEEN,CASE,CAST,FILTER,FROM,HAVING,%
    LIMIT,NOT,NULL,ON,ORDER,SELECT,USING,VALUE,VALUES,%
    WHEN,WHERE,AND,ASC,AVG,CHECK,COMMIT,COUNT,DECODE,DESC,DISTINCT,GROUP,IN,%
    LIKE,NUMBER,PREFIX,SUBSTR,SUM,%
    MIN,MAX,UNION,UPDATE,%
    ALL,ANY,CUBE,DEFAULT,DELETE,EXISTS,OR,%
    ROLE,ROLLUP,SET,SOME,TRIGGER,VIEW},%
  morendkeywords={BIT,BLOB,CHAR,CHARACTER,CLOB,DATE,DECIMAL,FLOAT,%
    INT,INTEGER,NUMERIC,SMALLINT,TIME,TIMESTAMP,VARCHAR},%
  sensitive=false,
  showstringspaces=false,
  morecomment=[l]--,%
  morecomment=[s]{/*}{*/},%
  morestring=[d]',%
  morestring=[d]"%
}[keywords,comments,strings]%

\lstdefinelanguage{XML}
{
  columns=fullflexible,
  alsoletter={-.:},
  showstringspaces=false,
  morecomment=[s]{!--}{--},
  morestring=[b]",
  morestring=[s]{>}{<},
  morecomment=[s]{<?}{?>},
}

\lstdefinelanguage{DTD}{
  alsoletter={-\#.:},
  keywords={,CDATA,DOCTYPE,ATTLIST,ELEMENT,EMPTY,ANY,ID,%
      IDREF,IDREFS,ENTITY,ENTITIES,NMTOKEN,NMTOKENS,NOTATION,%
      INCLUDE,IGNORE,SYSTEM,PUBLIC,NDATA,PUBLIC,%
      \#PCDATA,\#REQUIRED,\#IMPLIED,\#FIXED,}%
  showstringspaces=false,
  morecomment=[s]{<!--}{-->},
  morestring=[d]",
  morestring=[d]',
  stringstyle=\color{mygreen},
}

\lstdefinelanguage{BNF}
{
  columns=flexible,
  showstringspaces=false,
  morestring=[d]",
  morestring=[d]',
  morestring=[s]{\[}{\]},
  morestring=[s]{\#}{\ },
  stringstyle=\color{mygreen},
}

\lstdefinelanguage{TTL}
{
  columns=fullflexible,
  showstringspaces=false,
  morekeywords={@prefix,@base,@forSome,@forAll,@keywords},
  morecomment=[l]{\#},
  alsoletter={-?}, 
  basicstyle=\ttfamily\footnotesize, 
  morestring=[b][\color{black}]\",
}

\newfloat{lstfloat}{htbp}{lol}
\floatname{lstfloat}{Listing}



%% file: notation.tex
\newcommand{\iri}[1]{\texttt{\bfseries #1}}

\newcommand{\GAV}[2]{
\begin{minipage}[h]{0.4\linewidth}
\begin{lstlisting}[language=SQL,basicstyle=\ttfamily\footnotesize]
{#1}
\end{lstlisting}
\end{minipage}
\begin{minipage}[h]{0.1\linewidth}
  $\rightsquigarrow$
\end{minipage}
\begin{minipage}[h]{0.4\linewidth}
\begin{lstlisting}[language=TTL]
{#2}
\end{lstlisting}
\end{minipage}}

%% file: title.tex
R2RML Mappings in OBDA Systems:\\ Enabling Comparison among OBDA Tools

%% file: abstract.tex
In today's large enterprises there is a significant
increasing trend in the amount of data that has to be stored and processed.
To complicate this scenario the complexity of organizing and
managing a large collection of data, structured according to a single,
unified schema, makes so that there is almost never a single place where to look
to satisfy an information need.

The Ontology-Based Data Access (OBDA) paradigm aims at mitigating
this phenomenon by providing
to the users of the system a unified and shared conceptual view
of the domain of interest (ontology), while still enabling the data to
be stored in different data sources, which are managed
by a relational database.
In an OBDA system the link between the data stored at the sources
and the ontology is provided through a declarative specification
given in terms of a set of mappings.

In this work we focus on comparing two
of the available systems for OBDA, namely, Mastro and Ontop,
by adopting OBDA specifications based on W3C recommendations.
We first show how support for R2RML mappings has been integrated
in Mastro, which was the last feature missing in order to enable
the system to use specifications based solely on W3C recommendations
relevant to OBDA.
We then proceed in performing a comparison between these systems
over two OBDA specifications, the NPD Benchmark and the ACI
specification.

%% file: sec-intro.tex

\section{Introduction}
\label{sec:intro}
In today's large enterprises there is a significant
increasing trend in the amount of data that has to be stored and processed.
To complicate this scenario the complexity of organizing and
managing a large collection of data, structured according to a single,
unified schema, makes so that there is almost never a single place where to look
to satisfy an information need.

The Ontology-Based Data Access (OBDA) paradigm aims at providing
to the users of the system a unified and shared conceptual view
of the domain of interest (ontology), while still enabling the data to
be stored in different data sources, which are managed
by a relational database.
In an OBDA system the link between the data stored at the sources
and the ontology is provided through a declarative specification
given in terms of a set of mappings.

The interest in the adoption of the OBDA paradigm has lead
to the creation of prototype tools, that have evolved into full-fledged systems.
Although, while these tools effectively enable to answer queries over the ontology,
their use in industrial applications still represents a challenge
due to the performance requirements that have to be met.

Several studies on the performance comparison among tools that
enable semantic data integration have been performed, but
to the best of our knowledge none of them focused on comparing
fully-implemented systems designed to work in the OBDA setting,
in order to gain insight on the advantages and disadvantages of each
of the approach adopted, as this requires all the systems to be
able to use completely-standard specifications.

In this work we focus on comparing two systems
for OBDA, namely, Mastro\footnote{\url{https://www.obdasystems.com/mastro}}
and Ontop\footnote{\url{https://ontop.inf.unibz.it/}}.

In particular, we first enable the use of R2RML mappings in Mastro,
a tool for Ontology-Based Data Access developed at Sapienza, University of Rome.
R2RML is the W3C recommendation for expressing mappings in an OBDA specification,
and was the last feature missing in order to enable Mastro to use a
completely-standard specification.
We then proceed in performing a comparison between these systems
over two OBDA specifications, the NPD Benchmark and the ACI specification.

The rest of this work is organized as follows: In Section~\ref{sec:r2rml},
we briefly describe the basic notions of the R2RML mapping language, and
then show how such mappings can be interpreted for their use in Mastro.
In Section~\ref{sec:npd-benchmark}, we discuss the comparison over the
NPD Benchmark, a specification developed by the University of Oslo, and adapted
for its use as a benchmark in the OBDA setting\footnote{\url{https://github.com/ontop/npd-benchmark}}.
In Section~\ref{sec:aci}, we discuss the comparison over an
application of the OBDA paradigm, developed in collaboration between Sapienza
University of Rome and ACI Informatica\footnote{\url{http://www.informatica.aci.it/}},
that is used in a real industrial setting to evaluate the benefits of the OBDA approach.
Finally, in Section~\ref{sec:conclusions} we show some conclusions
and present some possible future works.


%% file: sec-r2rml.tex

\section{Adding R2RML Support to Mastro}
\label{sec:r2rml}

\emph{R2RML} (RDB to RDF mapping language) is the W3C recommendation
for expressing mappings from relational databases
to RDF datasets\cite{Das2012}. Such mappings provide the ability to view
existing relational data as RDF graphs, expressed in
a structure and target vocabulary of the mapping author's choice.

Compared to other RDB to RDF mapping languages R2RML can be classified
as a general purpose mapping language that allows to express
customized, domain-specific mappings.
An R2RML mapping is itself represented as an RDF graph.
A system that makes use of an R2RML mapping to provide access to an
RDF dataset from a relational database is called an R2RML \emph{processor}.
An R2RML processor could, for example, generate an RDF dump of the
relational data, or it could offer a SPARQL endpoint
over the \emph{virtual} RDF dataset represented by the database
and the R2RML mapping, through an interface that queries
the underlying database without explicitly materializing the dataset.
The latter scenario is particularly suitable in the case
of a system that wants to realize the OBDA paradigm
and is the reason why it as been considered as
a solution to represent the mapping between the ontology
and the data sources.

\begin{figure}[htbp]
  \centering
  \includegraphics[width=1.0\textwidth]{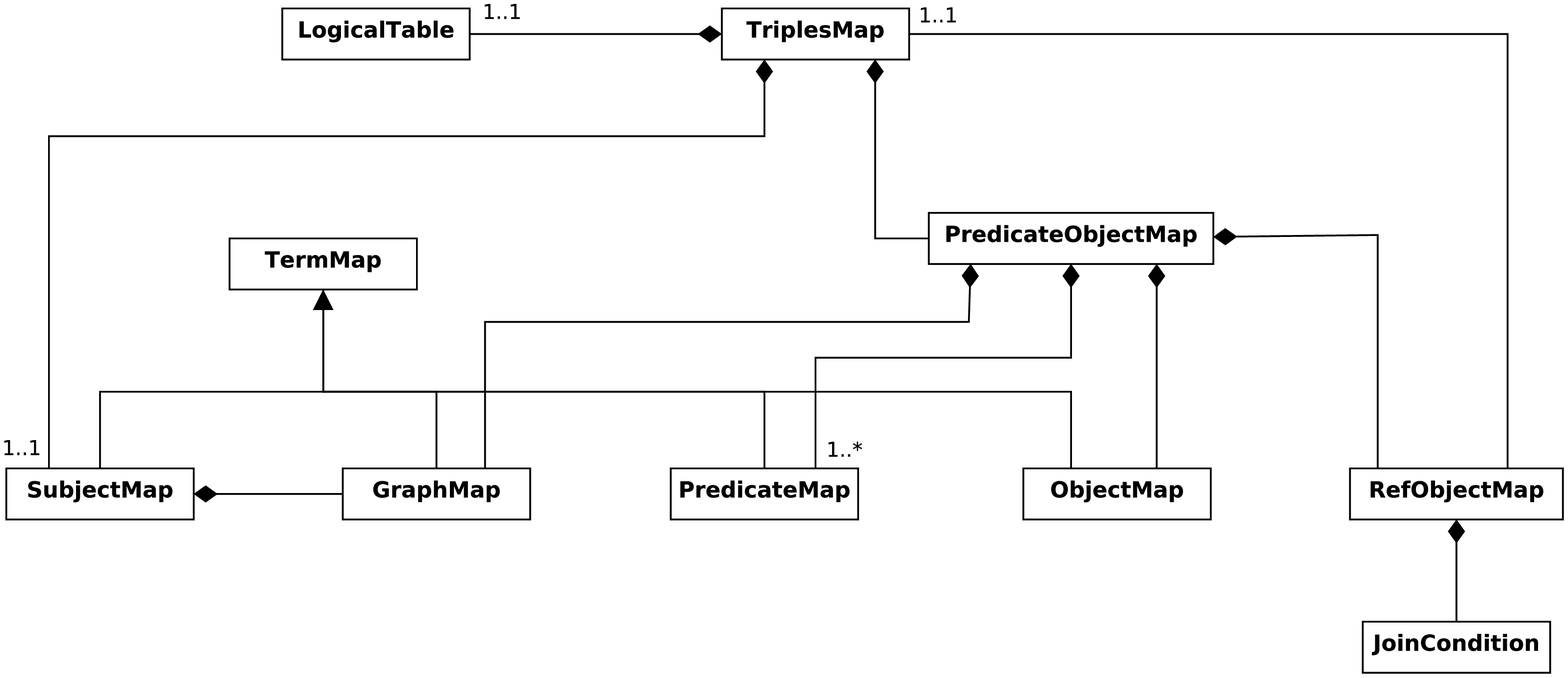}
  \caption{An overview of the R2RML classes}
  \label{fig:r2rml-overview}
\end{figure}

An overview of the main R2RML classes in the form of a UML class diagram
is provided in Figure~\ref{fig:r2rml-overview}.
Moreover, Listing~\ref{lst:r2rml-example} provides a complete example
of an R2RML mapping that maps the example database shown in
Figure~\ref{fig:r2rml-customers-example-db} to the RDF dataset
shown in Listing~\ref{lst:generated-rdf-example}.
For ease of presentation
we denote with \iri{rr:} the prefix for R2RML vocabulary terms,
with \iri{rdf:} the prefix for RDF vocabulary terms, and
\iri{ex:} the prefix for the specific target domain terms.

\begin{lstfloat}[H]
\begin{lstlisting}[language=TTL,numbers=left,frame=lines,basicstyle=\ttfamily\scriptsize]
@prefix rr: <http://www.w3.org/ns/r2rml#> .
@prefix ex: <http://data.example.com/> .

ex:CustomersMap rr:logicalTable [
        rr:sqlQuery "SELECT C_ID, C_NAME FROM customers" ;
        ] ;
    rr:subjectMap [
        rr:template "http://data.example.com/customer/{C_ID}" ;
        rr:class ex:Customer ;
        ] ;
    rr:predicateObjectMap [
        rr:predicate ex:name ;
        rr:objectMap [ rr:column "C_NAME" ]
        ] .

ex:ProductsMap rr:logicalTable [ rr:tableName "products" ] ;
    rr:subjectMap [
        rr:template "http://data.example.com/product/{P_ID}" ;
        rr:class ex:Product
        ] ;
    rr:predicateObjectMap [
        rr:predicate ex:price;
        rr:objectMap [ rr:column "P_PRICE" ]
        ] .

ex:OrdersMap rr:logicalTable [ rr:tableName "orders" ] ;
    rr:subjectMap [
        rr:template "http://data.example.com/order/{O_ID}" ;
        rr:class ex:Order ;
        ] ;
    rr:predicateObjectMap [
        rr:predicate ex:customer ;
        rr:objectMap [
            rr:parentTriplesMap ex:CustomersMap ;
            rr:joinCondition [
                rr:child "C_ID" ;
                rr:parent "C_ID" ;
                ]
            ] ;
    rr:predicateObjectMap [
    rr:predicate ex:product ;
        rr:objectMap [
            rr:parentTriplesMap ex:ProductsMap ;
            rr:joinCondition [
                rr:child "P_ID" ;
                rr:parent "P_ID" ;
                ]
            ]
        ] ;
    rr:predicateObjectMap [
        rr:predicate ex:quantity ;
        rr:objectMap [ rr:column "QUANTITY" ]
        ].
\end{lstlisting}
  \caption{Example of a complete R2RML mapping}
  \label{lst:r2rml-example}
\end{lstfloat}

\begin{lstfloat}[H]
\begin{lstlisting}[language=TTL,numbers=left,frame=lines,basicstyle=\ttfamily\scriptsize]
@prefix rdf: <http://www.w3.org/1999/02/22-rdf-syntax-ns#>
@prefix ex: <http://data.example.com/>

# RDF triples generated by ex:CustomersMap
<http://data.example.com/customer/3211> rdf:type ex:Customer .
<http://data.example.com/customer/3211> ex:name "Alice" .

<http://data.example.com/customer/3253> rdf:type ex:Customer .
<http://data.example.com/customer/3253> ex:name "Bob" .

# RDF triples generated by ex:ProductsMap
<http://data.example.com/product/2532> rdf:type ex:Product .
<http://data.example.com/product/2532> ex:price "12.00" .

<http://data.example.com/product/2533> rdf:type ex:Product .
<http://data.example.com/product/2533> ex:price "41.00" .

# RDF triples generated by ex:OrdersMap
<http://data.example.com/order/4301> rdf:type ex:Order ;
    ex:customer <http://data.example.com/customer/3211> ;
    ex:product <http://data.example.com/product/2532> ;
    ex:quantity "1" .

<http://data.example.com/orders/4302> rdf:type ex:Order ;
    ex:customer <http://data.example.com/customer/3211> ;
    ex:product <http://data.example.com/product/2533> ;
    ex:quantity "1" .

<http://data.example.com/orders/4303> rdf:type ex:Order ;
    ex:customer <http://data.example.com/customer/3253> ;
    ex:product <http://data.example.com/product/2532> ;
    ex:quantity "3" .
\end{lstlisting}
  \caption{Example of the generated RDF dataset}
  \label{lst:generated-rdf-example}
\end{lstfloat}

\vspace{2em}

\begin{figure}[hbtp]
  \centering
  \input{customersdb}

  \caption{Example relational database}
  \label{fig:r2rml-customers-example-db}
\end{figure}

\subsection{R2RML mappings in OBDA}
\label{sec:r2rml-obda}
Using Semantic Web technologies, the standard format for representing
knowledge is through RDF triples. RDB to RDF languages such
as R2RML allow to transform relational data into a set of RDF triples.

In OBDA the role of the mapping is to generate the
set of membership assertions that form the extensional
level of the ontology, and this can be represented
by a \emph{virtual} RDF graph when using these technologies.

We now show how an OBDA mapping can be expressed in R2RML,
and then proceed in illustrating how this can be integrated into the Mastro
system by showing a how a set of R2RML triples maps can be converted
into a set of mappings partitioned into view predicates and
ontology predicate mappings, and a set of mappings in the latter form
can be converted to a set of equivalent R2RML triples maps.

An OBDA mapping is represented by an expression of the form:

\[ \Phi(\vec{x}) \rightsquigarrow \Psi(\vec{y}, \vec{t}) \]
where $\Phi(\vec{x})$ is a conjunctive query over the source
schema, with free variables $\vec{x}$, $\vec{y} \subseteq \vec{x}$,
$\Psi(\vec{y}, \vec{t})$ is a conjunctive query over the ontology alphabet.

When trying to map the component of an OBDA mapping to the components
of an R2RML triples map, we face two problems: The first issue
is how to represent the query over the data source, that constitutes
the left-hand side of the mapping. The second issue is how to represent
the conjunction on the right-hand side, and how are the objects
build from the values returned by the logical table rows.

For what regards the first issue the most appropriate solution
is to use the effective SQL query of the logical table.
This corresponds exactly to the meaning of the query in the mapping assertion.

For what regards the second issue, if we apply the procedure for
generating RDF triples from a triples map, as defined by the standard,
we obtain a set of triples, all having the same subject,
corresponding to a conjunction of atoms composed by a unary atom
for each class specified in the subject map,
and a binary predicate for each of the predicate map-object map pair.

\pagebreak
\begin{exmp}
  \label{ex:objectmap}
Let's consider, for example, the triples map \iri{ex:CustomersMap}
of Listing~\ref{lst:r2rml-example}:

\vspace{1em}
\begin{lstlisting}[language=TTL,firstline=4,lastline=16,frame=lines,basicstyle=\ttfamily\scriptsize]
ex:CustomersMap rr:logicalTable [
        rr:sqlQuery "SELECT C_ID, C_NAME FROM customers" ;
        ] ;
    rr:subjectMap [
        rr:template "http://data.example.com/customer/{C_ID}" ;
        rr:class ex:Customer ;
        ] ;
    rr:predicateObjectMap [
        rr:predicate ex:name ;
        rr:objectMap [ rr:column "C_NAME" ]
        ] .

ex:ProductsMap rr:logicalTable [ rr:tableName "products" ] ;
\end{lstlisting}
\vspace{1em}

\noindent{}This triples map specifies that, for each logical table row
resulting from the evaluation of the SQL query:

\vspace{1em}
\begin{lstlisting}[language=SQL,basicstyle=\ttfamily\scriptsize]
SELECT C_ID, C_NAME FROM customers
\end{lstlisting}
\vspace{1em}

\noindent{}the system should produce the following set of assertions in the
virtual RDF graph:

\vspace{1em}
\begin{lstlisting}[language=TTL,basicstyle=\ttfamily\scriptsize]
ex:customer/{C_ID} rdf:type ex:Customer .
ex:customer/{C_ID} ex:name "{C_NAME}" .
\end{lstlisting}
\vspace{1em}
where \verb|{C_ID}| is replaced with the value of
the column \verb|C_ID|, and \verb|{C_NAME}| is replaced
with the value of the column \verb|C_NAME| in the logical table row.
This is equivalent to a mapping assertion of the form:

\vspace{1em}
\begin{minipage}[h]{0.4\linewidth}
\begin{lstlisting}[language=SQL,basicstyle=\ttfamily\scriptsize]
SELECT C_ID, C_NAME
FROM customers
\end{lstlisting}
\end{minipage}
\begin{minipage}[h]{0.1\linewidth}
  $\rightsquigarrow$
\end{minipage}
\begin{minipage}[h]{0.4\linewidth}
\begin{lstlisting}[language=TTL,basicstyle=\ttfamily\scriptsize]
ex:Customer(cust(C_ID)),
ex:name(cust(C_ID), C_NAME)
\end{lstlisting}
\end{minipage}
\vspace{1em}

\noindent{}where we replaced the IRIs that represent the subjects with
the function term $cust$. In fact, in R2RML, the role of the
constructors of the individuals in the ontology from the
values stored in the database is played by the
\emph{string templates}. This means that we can see
each different string template in the mapping as a different function symbol,
whose arity is the same as the number of placeholders in the template.

Additionally, in this setting, a mapping assertion of the previous form
has been shown\cite{PoggiLCGLR08} to be equivalent to the following set of mappings:

\pagebreak

\begin{framed}
\begin{minipage}[h]{0.4\linewidth}
\begin{lstlisting}[language=SQL,basicstyle=\ttfamily\scriptsize]
SELECT C_ID
FROM customers
\end{lstlisting}
\end{minipage}
\begin{minipage}[h]{0.1\linewidth}
  $\rightsquigarrow$
\end{minipage}
\begin{minipage}[h]{0.4\linewidth}
\begin{lstlisting}[language=TTL]
ex:Customer(cust(C_ID))
\end{lstlisting}
\end{minipage}

\begin{minipage}[h]{0.4\linewidth}
\begin{lstlisting}[language=SQL,basicstyle=\ttfamily\scriptsize]
SELECT C_ID, C_NAME
FROM customers
\end{lstlisting}
\end{minipage}
\begin{minipage}[h]{0.1\linewidth}
  $\rightsquigarrow$
\end{minipage}
\begin{minipage}[h]{0.4\linewidth}
\begin{lstlisting}[language=TTL]
ex:name(cust(C_ID), C_NAME)
\end{lstlisting}
\end{minipage}
\end{framed}
\end{exmp}


\begin{exmp}
  \label{ex:refobjectmap}
Let's now see the case of a triples map of the form corresponding
to the one of \iri{ex:OrdersMap} in the example of
Listing~\ref{lst:r2rml-example}:

\vspace{1em}
\begin{lstlisting}[language=TTL,firstline=28,lastline=55,frame=lines,basicstyle=\ttfamily\scriptsize]
        rr:template "http://data.example.com/order/{O_ID}" ;
        rr:class ex:Order ;
        ] ;
    rr:predicateObjectMap [
        rr:predicate ex:customer ;
        rr:objectMap [
            rr:parentTriplesMap ex:CustomersMap ;
            rr:joinCondition [
                rr:child "C_ID" ;
                rr:parent "C_ID" ;
                ]
            ] ;
    rr:predicateObjectMap [
    rr:predicate ex:product ;
        rr:objectMap [
            rr:parentTriplesMap ex:ProductsMap ;
            rr:joinCondition [
                rr:child "P_ID" ;
                rr:parent "P_ID" ;
                ]
            ]
        ] ;
    rr:predicateObjectMap [
        rr:predicate ex:quantity ;
        rr:objectMap [ rr:column "QUANTITY" ]
        ].
\end{lstlisting}
\vspace{1em}

\noindent{}
The logical table of this triples map specifies that the rows
to consider are directly those of the relation \verb|orders| in the
input database schema. Moreover, the presence of the referencing
object map imposes that, when generating the instances of
the roles \iri{ex:customer} and \iri{ex:product}, we need
to make sure that the objects we consider are only those
that are subjects of the assertions generated by the
\iri{ex:CustomersMap} and \iri{ex:ProductMap} respectively.
We do this by using the joint SQL query of the referencing
object map when generating the mappings. Similarly to the
previous case, the set of mappings corresponding to this
triples map are:

\pagebreak

\begin{framed}
\begin{minipage}[h]{0.4\linewidth}
\begin{lstlisting}[language=SQL,basicstyle=\ttfamily\scriptsize]
SELECT *
FROM orders
\end{lstlisting}
\end{minipage}
\begin{minipage}[h]{0.1\linewidth}
  $\rightsquigarrow$
\end{minipage}
\begin{minipage}[h]{0.4\linewidth}
\begin{lstlisting}[language=TTL]
ex:Order(ord(O_ID))
\end{lstlisting}
\end{minipage}

\begin{minipage}[h]{0.4\linewidth}
\begin{lstlisting}[language=SQL,basicstyle=\ttfamily\scriptsize]
SELECT *
FROM
  (SELECT C_ID, C_NAME
   FROM customers) AS parent,
  (SELECT *
   FROM orders) AS child
WHERE child.C_ID=parent.C.ID
\end{lstlisting}
\end{minipage}
\begin{minipage}[h]{0.05\linewidth}
  $\rightsquigarrow$
\end{minipage}
\begin{minipage}[h]{0.3\linewidth}
\begin{lstlisting}[language=TTL]
ex:customer(ord(O_ID), cust(C_ID))
\end{lstlisting}
\end{minipage}

\begin{minipage}[h]{0.4\linewidth}
\begin{lstlisting}[language=SQL,basicstyle=\ttfamily\scriptsize]
SELECT *
FROM
  (SELECT *
   FROM products) AS parent,
  (SELECT *
   FROM orders) AS child
WHERE child.P_ID=parent.P_ID
\end{lstlisting}
\end{minipage}
\begin{minipage}[h]{0.05\linewidth}
  $\rightsquigarrow$
\end{minipage}
\begin{minipage}[h]{0.3\linewidth}
\begin{lstlisting}[language=TTL]
ex:product(ord(O_ID), prod(C_ID))
\end{lstlisting}
\end{minipage}
\end{framed}
where we introduced two new function terms $ord$ and $prod$,
corresponding to the string templates of the customers
and products respectively.
\end{exmp}

\subsection{Importing R2RML Mappings in Mastro}
Up to now we have seen how a set of R2RML mappings
are interpreted as a form of mappings that can be used in
an OBDA system to generate the virtual ABox assertions.

In Mastro, the set of mappings is composed by a pair
$\mathcal{M} = \langle \mathcal{M}_{v},\mathcal{M}_{o} \rangle$, where:
\begin{itemize}
\item $\mathcal{M}_{v}$ is a set of assertions of the form:
  \[ q_{DB}(\vec{x}) \rightsquigarrow v(\vec{x}) \]
  where $q_{DB}$ is an SQL query, and $v$ is a view predicate.
\item $\mathcal{M}_{o}$ is a set of assertions of the form:
  \[ q_{v}(\vec{x}) \rightsquigarrow P\left(\vec{x}, \vec{t}\right) \]
  where $P$ is an atomic ontology predicate, built from function
  terms in $\vec{t}$ applied over the variables in $\vec{x}$.
\end{itemize}

The view predicates act as an intermediate level of abstraction and
have a dual purpose: On one side they enable the database manager
to focus on improving the efficiency of the database management
system, by having a description of the relevant queries that will
have to be performed. On the other side they free the ontology
designer from having to consider the technical detail of the database schema.

In R2RML the closest notion to the views used in Mastro are the
logical tables. Unfortunately, the R2RML language does not provide
the vocabulary for expressing a logical table as an arbitrary conjunction
of other logical tables (except when using referencing object maps,
which are restricted to express foreign key relationships
among logical tables), but relies completely on the queries
at the SQL level. This does not really pose a problem
when importing an R2RML mapping, as we can see each logical table
as a view mapping, where the associated query is the effective
SQL query of the logical table, and build the ontology mappings
for each of the assertions generated by the triples map using
this newly generated view mapping. The challenge is how to
encode mappings that are expressed as arbitrary conjunction of
view predicates in R2RML.
The solution we adopt in order to capture the full semantics
of the mappings in Mastro is to create a new logical table
for each of this conjunctions, where the corresponding SQL
query is the unfolding of the query over the views.
In the following, we illustrate both processes by means
of some examples.

\begin{exmp}
  \label{ex:mastro-objectmap}
Starting from the example~\ref{ex:objectmap} from the previous section,
in order to obtain a set of mappings in the form used by Mastro,
we need to introduce an auxiliary
view predicate, corresponding to the SQL query of the
left-hand side of the original mapping, and substitute
it in the left-hand side of the previously shown mappings:

\begin{framed}
  \begin{itemize}
  \item $M_{v}$:\\
    \begin{minipage}[h]{0.4\linewidth}
\begin{lstlisting}[language=SQL,basicstyle=\ttfamily\scriptsize]
SELECT C_ID, C_NAME
FROM customers
\end{lstlisting}
    \end{minipage}
    \begin{minipage}[h]{0.05\linewidth}
      $\rightsquigarrow$
    \end{minipage}
    \begin{minipage}[h]{0.4\linewidth}
\begin{lstlisting}[language=TTL]
customers_view(C_ID, C_NAME)
\end{lstlisting}
    \end{minipage}

  \item $M_{o}$:\\
    \begin{minipage}[h]{0.5\linewidth}
\begin{lstlisting}[language=SQL,basicstyle=\ttfamily\scriptsize]
customers_view(C_ID, C_NAME)
\end{lstlisting}
    \end{minipage}
    \begin{minipage}[h]{0.05\linewidth}
      $\rightsquigarrow$
    \end{minipage}
    \begin{minipage}[h]{0.4\linewidth}
\begin{lstlisting}[language=TTL]
ex:Customer(cust(C_ID))
\end{lstlisting}
    \end{minipage}

    \begin{minipage}[h]{0.5\linewidth}
\begin{lstlisting}[language=SQL,basicstyle=\ttfamily\scriptsize]
customers_view(C_ID, C_NAME)
\end{lstlisting}
    \end{minipage}
    \begin{minipage}[h]{0.05\linewidth}
      $\rightsquigarrow$
    \end{minipage}
    \begin{minipage}[h]{0.4\linewidth}
\begin{lstlisting}[language=TTL]
ex:name(cust(C_ID), C_NAME)
\end{lstlisting}
    \end{minipage}
  \end{itemize}
\end{framed}
\end{exmp}

\begin{exmp}[Importing a mapping from a referencing object map]
  \label{ex:mastro-refobjectmap}
When translating R2RML mappings of the type shown in
example~\ref{ex:refobjectmap} we can express the query of the referencing
object map directly as queries over the view predicates:

\pagebreak

\begin{framed}
  \begin{itemize}
  \item $M_{v}$:\\
    \begin{minipage}[h]{0.4\linewidth}
\begin{lstlisting}[language=SQL,basicstyle=\ttfamily\scriptsize]
SELECT C_ID, C_NAME
FROM customers
\end{lstlisting}
    \end{minipage}
    \begin{minipage}[h]{0.1\linewidth}
      $\rightsquigarrow$
    \end{minipage}
    \begin{minipage}[h]{0.4\linewidth}
\begin{lstlisting}[language=TTL,basicstyle=\ttfamily\scriptsize]
customers_view(C_ID, C_NAME)
\end{lstlisting}
    \end{minipage}

    \begin{minipage}[h]{0.4\linewidth}
\begin{lstlisting}[language=SQL,basicstyle=\ttfamily\scriptsize]
SELECT P_ID, P_PRICE
FROM products
\end{lstlisting}
    \end{minipage}
    \begin{minipage}[h]{0.05\linewidth}
      $\rightsquigarrow$
    \end{minipage}
    \begin{minipage}[h]{0.4\linewidth}
\begin{lstlisting}[language=TTL,basicstyle=\ttfamily\scriptsize]
products_view(P_ID, P_PRICE)
\end{lstlisting}
    \end{minipage}

    \begin{minipage}[h]{0.4\linewidth}
\begin{lstlisting}[language=SQL,basicstyle=\ttfamily\scriptsize]
SELECT O_ID, C_ID,
       P_ID, QUANTITY
FROM orders
\end{lstlisting}
    \end{minipage}
    \begin{minipage}[h]{0.05\linewidth}
      $\rightsquigarrow$
    \end{minipage}
    \begin{minipage}[h]{0.4\linewidth}
\begin{lstlisting}[language=TTL,basicstyle=\ttfamily\scriptsize]
orders_view(O_ID, C_ID, P_ID, QUANTITY)
\end{lstlisting}
    \end{minipage}

  \item $M_{o}$:\\
    \begin{minipage}[h]{0.5\linewidth}
\begin{lstlisting}[language=SQL,basicstyle=\ttfamily\scriptsize]
  orders_view(O_ID, C_ID,
              P_ID, QUANTITY)
\end{lstlisting}
    \end{minipage}
    \begin{minipage}[h]{0.05\linewidth}
      $\rightsquigarrow$
    \end{minipage}
    \begin{minipage}[h]{0.4\linewidth}
\begin{lstlisting}[language=TTL,basicstyle=\ttfamily\scriptsize]
ex:Order(ord(C_ID))
\end{lstlisting}
    \end{minipage}

    \begin{minipage}[h]{0.5\linewidth}
\begin{lstlisting}[language=SQL,basicstyle=\ttfamily\scriptsize]
customers_view(C_ID,C_NAME)
orders_view(O_ID,C_ID,
            P_ID,QUANTITY)
\end{lstlisting}
    \end{minipage}
    \begin{minipage}[h]{0.05\linewidth}
      $\rightsquigarrow$
    \end{minipage}
    \begin{minipage}[h]{0.4\linewidth}
\begin{lstlisting}[language=TTL,basicstyle=\ttfamily\scriptsize]
ex:customer(ord(O_ID), cust(C_ID))
\end{lstlisting}
    \end{minipage}

    \begin{minipage}[h]{0.5\linewidth}
\begin{lstlisting}[language=SQL,basicstyle=\ttfamily\scriptsize]
products_view(P_ID,P_PRICE)
orders_view(O_ID,C_ID,
            P_ID,QUANTITY)
\end{lstlisting}
    \end{minipage}
    \begin{minipage}[h]{0.05\linewidth}
      $\rightsquigarrow$
    \end{minipage}
    \begin{minipage}[h]{0.4\linewidth}
\begin{lstlisting}[language=TTL,basicstyle=\ttfamily\scriptsize]
ex:product(ord(O_ID), prod(P_ID))
\end{lstlisting}
    \end{minipage}
  \end{itemize}
\end{framed}
\noindent
where the join between the parent and child triples map is
performed on the left-hand side of the mappings in $M_{o}$
by performing unification over the attributes stated
in the join conditions.
\end{exmp}

\subsection{Exporting Mastro Mappings in R2RML}
\label{sec:mastro-r2rml-export}
The procedure for exporting a set of mappings expressed in the Mastro
mapping format is the dual of the one previously described,
with an important distinction:
We need to introduce a new logical table for each of the
conjunctions on the left-hand side of the ontology predicate mappings
in order to be able to capture in R2RML the complete semantics
that form the left-hand side of the mapping,
This is because R2RML does not allow to express logical tables
as arbitrary conjunctive queries of other logical tables.
To obtain the SQL query of the generated logical table
we unfold the query over the view predicates.
During this translation all the constraints over the views, and the
other optimizations expressed in the internal mapping format
have to be discarded, as they cannot be expressed in R2RML.

\begin{exmp}[Exporting a mapping with a single view atom]
  Consider the case shown in example~\ref{ex:mastro-objectmap}.
  In this case it is enough to generate a logical table for the view
  \verb|customers_view|, and a triples map for each of the mappings
  corresponding to the right-hand side of the ontology predicate
  mappings:

  \vspace{1em}
\begin{lstlisting}[language=TTL,frame=lines,basicstyle=\ttfamily\scriptsize]
_:customer_view rr:sqlQuery " SELECT C_ID, C_NAME FROM customers "
ex:Customer rr:logicalTable _:customer_view
            rr:subjectMap [
                rr:template "http://data.example.com/customer/{C_ID}" ;
                rr:class ex:Customer
            ]

ex:name rr:logicalTable _:customer_view
        rr:subjectMap [
            rr:template "http://data.example.com/customer/{C_ID}"
        ]
        rr:predicateObjectMap [
            rr:predicate ex:name ;
            rr:objectMap [ rr:column "C_NAME" ]
        ]
\end{lstlisting}
\end{exmp}

\begin{exmp}[Exporting a mapping with a conjunction of view atoms]
Now consider the mappings shown in example~\ref{ex:mastro-refobjectmap}.
The mappings for the roles \iri{ex:customer} and \iri{ex:products}
are expressed as a conjunction of view predicates. Even though
there may be cases where this type of assertion can be translated
with the use of a referencing object map, we cannot assume that
this is always the case. A general solution, that preserves the
semantics of the mapping, is to transform the conjunction on the left-hand
side of both roles into a new logical table, corresponding to the
unfolding of the query over the views. This means generating
triples maps in the following form:

\vspace{1em}
\begin{lstlisting}[language=TTL,frame=lines,basicstyle=\ttfamily\scriptsize]
ex:customerMap rr:logicalTable [
                   rr:sqlQuery """
                               SELECT v1.C_ID AS C_ID, v2.O_ID AS O_ID
                               FROM
                                 (SELECT C_ID, C_NAME
                                  FROM customers) AS v1,
                                 (SELECT O_ID, C_ID
                                  FROM orders) AS v2
                               WHERE v1.C_ID=v2.C.ID
                               """.
               ] ;
             rr:subjectMap [
                 rr:template "http://data.example.com/order/{O_ID}" ;
             ] ;
             rr:predicateObjectMap [
                 rr:predicate ex:customer ;
                 rr:objectMap [
                     rr:template "http://data.example.com/customer/{C_ID}"
                 ]
             ] .
\end{lstlisting}
\end{exmp}


%% file: customersdb.tex

\begin{minipage}{0.55\linewidth}
  \underline{\textbf{Customers}}
\end{minipage}
\begin{minipage}{0.35\linewidth}
  \underline{\textbf{Products}}
\end{minipage}

\vspace{0.5em}
\begin{minipage}{0.55\linewidth}
  \begin{tabular}{|c|c|c|}
    \hline
    \underline{\textbf{C\_ID}} & \textbf{C\_NAME} \\
    \hline
    3211                       & Alice            \\
    \hline
    3253                       & Bob              \\
    \hline
  \end{tabular}
\end{minipage}
\begin{minipage}{0.35\linewidth}
  \begin{tabular}{|c|c|c|}
    \hline
    \underline{\textbf{P\_ID}} & \textbf{P\_PRICE} \\
    \hline
    2532                       & 12.00             \\
    \hline
    2533                       & 41.00             \\
    \hline
  \end{tabular}
\end{minipage}

\vspace{1.0em}
\begin{minipage}{0.4\linewidth}

\end{minipage}
\begin{minipage}{0.5\linewidth}
  \underline{\textbf{Orders}}
\end{minipage}

\vspace{0.5em}
\begin{minipage}{1.0\linewidth}
  \centering
  \begin{tabular}{|c|c|c|c|}
    \hline
    \underline{\textbf{O\_ID}} & \textbf{C\_ID} & \textbf{P\_ID} & \textbf{QUANTITY} \\
    \hline
    4301                       & 3211           & 2532           & 1                 \\
    \hline
    4302                       & 3211           & 2533           & 1                 \\
    \hline
    4303                       & 3253           & 2532           & 3                 \\
    \hline
  \end{tabular}
\end{minipage}

%% file: sec-npd-prep.tex

\section{Comparison on the NPD Benchmark}
\label{sec:npd-benchmark}
The increase in the rate of adoption of the OBDA paradigm has lead
to the creation of several prototype tools. Although, while these
tools effectively enable to answer queries over the ontology,
their use in industrial applications still represents a challenge
due to the performance requirements that have to be met.

Towards this direction, several benchmarks from the Semantic Web world have
been proposed as a way of evaluating the performance of an OBDA system
\cite{GuoPH05,BizerS09}.

Unfortunately, all these benchmarks lack a fundamental property,
that is the presence of a complex and expressive ontology,
which is required in order to effectively evaluate the performance of
an OBDA system in an industrial application setting.
For this purpose, recently a new benchmark has been proposed,
based on the \emph{Norvegian Pretroleum Directorate (NPD) FactPages}~\footnote{\url{http://factpages.npd.no/factpages/}}.

The Norwegian Petroleum Directorate is a governmental organization
whose main objective is to contribute to maximize
the value that society can obtain from the oil and gas activities
\cite{LantiRXC15}.
The NPD FactPages contains information regarding the petroleum
activities on the Norwegian continental shelf. Such information is
actively used by oil companies like Statoil. The Factpages are periodically
synchronized with the NPD’s databases.

The NPD ontology has been mapped to the NPD FactPages and stored
in a relational database. Together with the ontology, the
benchmark is provided with a dump of the original database
created from the NPD FactPages, the set of mappings expressed in
the R2RML mapping language, and a set of queries that have
been formulated by domain experts starting from an informal set of questions
provided to the users of the NPD FactPages.

In the following we will first describe the main characteristics
of what composes the NPD
Benchmark\footnote{\url{https://github.com/ontop/npd-benchmark}}
specification, with respect to version 1.9.
We will then proceed in illustrating
the setup for our experimentation, with a presentation of the
results obtained.

\paragraph{Ontology}
The NPD Ontology~\cite{SkjaevelandLH13} describes
activities on the Norwegian continental shelf (NCS), e.g., about companies
that own or operate petroleum fields, results of tests taken during drilling,
geographical data for physical installations and the areas of fields
and seismic surveys, transfers of shares of fields between companies,
and production results measured in volumes of petroleum
\cite{SkjaevelandLH13}. The ontology has been
created by the University of Oslo, and presents rich hierarchies of classes and
properties, axioms that infer new objects, and disjointness assertions.

The ontology is specified in OWL and for the purpose of the benchmark
it has been restricted to the fragment corresponding to the OWL 2 QL
profile. Overall is composed by about $350$ concepts, $142$ roles
and $238$ attributes, with a maximum hierarchy depth of $10$.
This restriction is essential for its use in the context of OBDA
as guarantees first-order rewritability for the class of
union of conjunctive queries.

\paragraph{Mappings}
The NPD specification provides a set of $1173$ mapping assertions,
characterized by an average of $1.7$ joins per query.
The mappings have been partially bootstrapped automatically from the database
and the ontology, and partially created by hand,
and are specified in the R2RML mapping language.

The mappings have purposely not optimized, to measure the efficiency
of the optimization strategies employed by an OBDA system.
This means that the number of mappings that refer to the same ontology
predicate is in general very large, up to about $30$ in some cases.

\paragraph{Query Set}
The latest revision of the NPD benchmark devises $30$ between real world
and technical queries of different complexity, expressed in SPARQL,
and defined by domain experts
starting from an informal set of questions to the users of the NPD
FactPages. Among the set of queries, some have been specifically
generated to stress the efficiency of a system when reasoning
with respect to existential variables.

Some of the characteristics of the queries are the presence of concepts
with a rich hierarchy and the presence of aggregations.
For the purpose of this experimentation, we are only interested in
the subset of these SPARQL queries corresponding to the class of
union of conjunctive queries, as this is the semantics
for SPARQL queries that is adopted by Mastro.
The only exception is that we make is for the use of duplicate elimination
from the results. This requires to changes part of the query set.

In the following we present each individual query,
together with its \emph{CQ} restriction, the changes
required, and a brief description of the query.

%

\begin{lstfloat}[H]
\bgroup 
\singlespacing
\begin{tcolorbox}

{\bfseries\large NPD query q1}
\begin{framed}
\begin{lstlisting}[language=SPARQL,basicstyle=\ttfamily\scriptsize]
SELECT DISTINCT ?licenceURI ?interest ?date
WHERE {
  ?licenceURI a npdv:ProductionLicence .
  
  [ ] a npdv:ProductionLicenceLicensee ;
  npdv:dateLicenseeValidFrom ?date ;
  npdv:licenseeInterest ?interest ;
  npdv:licenseeForLicence ?licenceURI .   
  FILTER(?date > "1979-12-31T00:00:00"^^xsd:dateTime)	
}
\end{lstlisting}
\end{framed}

{\bfseries\large CQ query q1}
\begin{framed}
\begin{lstlisting}[language=SPARQL,basicstyle=\ttfamily\scriptsize]
SELECT DISTINCT ?licenceURI ?interest ?date
WHERE {
  ?licenceURI a npdv:ProductionLicence .
  
  [ ] a npdv:ProductionLicenceLicensee ;
  npdv:dateLicenseeValidFrom ?date ;
  npdv:licenseeInterest ?interest ;
  npdv:licenseeForLicence ?licenceURI .   
  FILTER(?date > "1979-12-31T00:00:00"^^xsd:dateTime)	
}
\end{lstlisting}
\end{framed}

{\bfseries\large Modifications}
\begin{framed}
No modification is required for query $q1$.
\end{framed}

{\bfseries\large Description}
\begin{framed}
  Query q1 asks for the licenses, the interest applied to the respective
  licensees, and from when they are valid.
\end{framed}

\end{tcolorbox}
\egroup 
\caption{NPD Query q1}
\label{lst:npd-q1}
\end{lstfloat}

\pagebreak

\begin{lstfloat}[H]
\bgroup 
\singlespacing
\begin{tcolorbox}

{\bfseries\large NPD query q2}
\begin{framed}
\begin{lstlisting}[language=SPARQL,basicstyle=\ttfamily\scriptsize]
SELECT ?licenceURI ?company ?date 
WHERE {
  ?licenceURI a npdv:ProductionLicence .
  
  [ ] a npdv:ProductionLicenceOperator ;
  npdv:dateOperatorValidFrom ?date ;
  npdv:licenceOperatorCompany [ npdv:name ?company ] ;
  npdv:operatorForLicence ?licenceURI .

  FILTER(?date > "1979-12-31T00:00:00"^^xsd:dateTime)
  
} ORDER BY ?licenceURI
\end{lstlisting}
\end{framed}

{\bfseries\large CQ query q2}
\begin{framed}
\begin{lstlisting}[language=SPARQL,basicstyle=\ttfamily\scriptsize]
SELECT ?licenceURI ?company ?date 
WHERE {
  ?licenceURI a npdv:ProductionLicence .
  
  [ ] a npdv:ProductionLicenceOperator ;
  npdv:dateOperatorValidFrom ?date ;
  npdv:licenceOperatorCompany [ npdv:name ?company ] ;
  npdv:operatorForLicence ?licenceURI .

  FILTER(?date > "1979-12-31T00:00:00"^^xsd:dateTime)  
}
\end{lstlisting}
\end{framed}

{\bfseries\large Modifications}
\begin{framed}
  The CQ version of query q2 is obtained by removing the
  {\bfseries ORDER BY} clause.
\end{framed}

{\bfseries\large Description}
\begin{framed}
  Query $q2$ asks for the operators for licences whose contracts
  were started after $1980$.
\end{framed}
\end{tcolorbox}
\egroup 
\caption{NPD Query q2}
\label{lst:npd-q2}
\end{lstfloat}

\pagebreak

\begin{lstfloat}[H]
\bgroup 
\singlespacing
\begin{tcolorbox}

{\bfseries\large NPD query q3}
\begin{framed}
\begin{lstlisting}[language=SPARQL,basicstyle=\ttfamily\scriptsize]
SELECT ?licence ?dateGranted ?dateValidTo
WHERE {
  [ ] a npdv:ProductionLicence ;
  npdv:name ?licence ;
  npdv:dateLicenceGranted ?dateGranted ;
  npdv:dateLicenceValidTo ?dateValidTo .

  FILTER(?dateValidTo > "1979-12-31T00:00:00"^^xsd:dateTime)
} ORDER BY ?licence
\end{lstlisting}
\end{framed}

{\bfseries\large CQ query q3}
\begin{framed}
\begin{lstlisting}[language=SPARQL,basicstyle=\ttfamily\scriptsize]
SELECT ?licence ?dateGranted ?dateValidTo
WHERE {
  [ ] a npdv:ProductionLicence ;
  npdv:name ?licence ;
  npdv:dateLicenceGranted ?dateGranted ;
  npdv:dateLicenceValidTo ?dateValidTo .

  FILTER(?dateValidTo > "1979-12-31T00:00:00"^^xsd:dateTime)
}
\end{lstlisting}
\end{framed}

{\bfseries\large Modifications}
\begin{framed}
  The CQ version of query $q3$ is obtained by removing the
  {\bfseries ORDER BY} clause with respect to the ?licence variable.
\end{framed}

{\bfseries\large Description}
\begin{framed}
Query $q3$ asks for the licences whose expiration dates were after $1980$.
\end{framed}
\end{tcolorbox}
\egroup 
\caption{NPD Query q3}
\label{lst:npd-q3}
\end{lstfloat}

\pagebreak

\begin{lstfloat}[H]
\bgroup 
\singlespacing
\begin{tcolorbox}

{\bfseries\large NPD query q4}
\begin{framed}
\begin{lstlisting}[language=SPARQL,basicstyle=\ttfamily\scriptsize]
SELECT ?licence ?company ?licenseeFrom
WHERE {
  [ ] npdv:licenseeForLicence 
  [ a npdv:ProductionLicence ;
    npdv:name ?licence ] ;
  npdv:licenceLicensee [ npdv:name ?company ] ;
  npdv:dateLicenseeValidFrom ?licenseeFrom .
  
  FILTER(?licenseeFrom > "1979-12-31T00:00:00"^^xsd:dateTime)
} ORDER BY ?licence ASC(?licenseeFrom)
\end{lstlisting}
\end{framed}

{\bfseries\large CQ query q4}
\begin{framed}
\begin{lstlisting}[language=SPARQL,basicstyle=\ttfamily\scriptsize]
SELECT ?licence ?company ?licenseeFrom
WHERE {
  [ ] npdv:licenseeForLicence 
  [ a npdv:ProductionLicence ;
    npdv:name ?licence ] ;
  npdv:licenceLicensee [ npdv:name ?company ] ;
  npdv:dateLicenseeValidFrom ?licenseeFrom .
  
  FILTER(?licenseeFrom > "1979-12-31T00:00:00"^^xsd:dateTime)
}
\end{lstlisting}
\end{framed}

{\bfseries\large Modifications}
\begin{framed}
  The CQ of query $q4$ is obtained by removing the {\bfseries ORDER BY}
  clause.
\end{framed}

\end{tcolorbox}
\egroup 
\caption{NPD Query q4}
\label{lst:npd-q4}
\end{lstfloat}

\pagebreak

\begin{lstfloat}[H]
\bgroup 
\singlespacing
\begin{tcolorbox}

{\bfseries\large NPD query q5}
\begin{framed}
\begin{lstlisting}[language=SPARQL,basicstyle=\ttfamily\scriptsize]
SELECT  ?fr ?OE ?oil ?gas ?NGL ?con  
WHERE {
  ?fr a npdv:FieldReserve ;
  npdv:remainingCondensate     ?con ;
  npdv:remainingGas            ?gas ;
  npdv:remainingNGL            ?NGL ;
  npdv:remainingOil            ?oil ;
  npdv:remainingOilEquivalents ?OE  .
  
  FILTER(?gas < 100)
} ORDER BY DESC(?OE)
\end{lstlisting}
\end{framed}

{\bfseries\large CQ query q5}
\begin{framed}
\begin{lstlisting}[language=SPARQL,basicstyle=\ttfamily\scriptsize]
SELECT  ?fr ?OE ?oil ?gas ?NGL ?con  
WHERE {
  ?fr a npdv:FieldReserve ;
  npdv:remainingCondensate     ?con ;
  npdv:remainingGas            ?gas ;
  npdv:remainingNGL            ?NGL ;
  npdv:remainingOil            ?oil ;
  npdv:remainingOilEquivalents ?OE  .
  
  FILTER(?gas < 100)
}
\end{lstlisting}
\end{framed}

{\bfseries\large Modifications}
\begin{framed}
  The CQ of query $q5$ is obtained by removing the {\bfseries ORDER BY}
  clause.
\end{framed}

\end{tcolorbox}
\egroup 
\caption{NPD Query q5}
\label{lst:npd-q5}
\end{lstfloat}

\pagebreak

\begin{lstfloat}[H]
\bgroup 
\singlespacing
\begin{tcolorbox}

{\bfseries\large NPD query q6}
\begin{framed}
\begin{lstlisting}[language=SPARQL,basicstyle=\ttfamily\scriptsize]
SELECT DISTINCT ?wellbore (?length AS ?lenghtM) ?company ?year 
WHERE {
  ?wc npdv:coreForWellbore
  [ rdf:type                      npdv:Wellbore ;
    npdv:name                     ?wellbore ;
    npdv:wellboreCompletionYear   ?year ;
    npdv:drillingOperatorCompany  [ npdv:name ?company ] 
  ] .
  { ?wc npdv:coresTotalLength ?length } 

  FILTER(?year >= "2008"^^xsd:integer &&
         ?length > 50)
} ORDER BY ?wellbore
\end{lstlisting}
\end{framed}

{\bfseries\large Description}
\begin{framed}
  This is a query that asks for the wellbores, their length,
  and the companies that completed the drilling of the wellbore after 2008,
  and sampled more than 50m of cores. The use of graph patterns of this
  form is not supported in Mastro, and so it has not been considered.
\end{framed}
\end{tcolorbox}
\egroup 
\caption{NPD Query q6}
\label{lst:npd-q6}
\end{lstfloat}

\pagebreak

\begin{lstfloat}[H]
\bgroup 
\singlespacing
\begin{tcolorbox}

{\bfseries\large NPD query q7}
\begin{framed}
\begin{lstlisting}[language=SPARQL,basicstyle=\ttfamily\scriptsize]
SELECT *
WHERE {
  [ ] a npdv:FieldMonthlyProduction ;
  npdv:productionYear         ?year;
  npdv:productionMonth        ?month;
  npdv:producedCondensate     ?con ;
  npdv:producedGas            ?gas ;
  npdv:producedNGL            ?NGL ;
  npdv:producedOil            ?oil ;
  npdv:producedOilEquivalents ?maxOE  .

  FILTER(?gas < 100)
} 
\end{lstlisting}
\end{framed}

{\bfseries\large CQ query q7}
\begin{framed}
\begin{lstlisting}[language=SPARQL,basicstyle=\ttfamily\scriptsize]
SELECT *
WHERE {
  [ ] a npdv:FieldMonthlyProduction ;
  npdv:productionYear         ?year;
  npdv:productionMonth        ?month;
  npdv:producedCondensate     ?con ;
  npdv:producedGas            ?gas ;
  npdv:producedNGL            ?NGL ;
  npdv:producedOil            ?oil ;
  npdv:producedOilEquivalents ?maxOE  .

  FILTER(?gas < 100)
} 
\end{lstlisting}
\end{framed}

{\bfseries\large Modifications}
\begin{framed}
  No modification is required for query $q7$.
\end{framed}

\end{tcolorbox}
\egroup 
\caption{NPD Query q7}
\label{lst:npd-q7}
\end{lstfloat}

\pagebreak

\begin{lstfloat}[H]
\bgroup 
\singlespacing
\begin{tcolorbox}

{\bfseries\large NPD query q8}
\begin{framed}
\begin{lstlisting}[language=SPARQL,basicstyle=\ttfamily\scriptsize]
SELECT *
WHERE {
  [ npdv:productionYear ?year ;
    npdv:productionMonth ?m ;
    npdv:producedGas     ?g ;
    npdv:producedOil     ?o 
  ]
  FILTER (?year > 1999) 
  FILTER(?m >= 1 && ?m <= 6 )
} 
\end{lstlisting}
\end{framed}

{\bfseries\large CQ query q8}
\begin{framed}
\begin{lstlisting}[language=SPARQL,basicstyle=\ttfamily\scriptsize]
SELECT *
WHERE {
  [ npdv:productionYear ?year ;
    npdv:productionMonth ?m ;
    npdv:producedGas     ?g ;
    npdv:producedOil     ?o 
  ]
  FILTER (?year > 1999) 
  FILTER(?m >= 1 && ?m <= 6 )
} 
\end{lstlisting}
\end{framed}

{\bfseries\large Modifications}
\begin{framed}
  No modification is required for query $q8$.
\end{framed}

\end{tcolorbox}
\egroup 
\caption{NPD Query q8}
\label{lst:npd-q8}
\end{lstfloat}

\pagebreak

\begin{lstfloat}[H]
\bgroup 
\singlespacing
\begin{tcolorbox}

{\bfseries\large NPD query q9}
\begin{framed}
\begin{lstlisting}[language=SPARQL,basicstyle=\ttfamily\scriptsize]
SELECT *
WHERE { 
  [ ] a npdv:Facility ;
  npdv:name ?facility ;
  npdv:registeredInCountry ?country;
  npdv:idNPD ?id . 
  FILTER (?id > "400000"^^xsd:integer)
} ORDER BY ?facility

\end{lstlisting}
\end{framed}

{\bfseries\large CQ query q9}
\begin{framed}
\begin{lstlisting}[language=SPARQL,basicstyle=\ttfamily\scriptsize]
SELECT *
WHERE { 
  [ ] a npdv:Facility ;
  npdv:name ?facility ;
  npdv:registeredInCountry ?country;
  npdv:idNPD ?id . 
  FILTER (?id > "400000"^^xsd:integer)
}

\end{lstlisting}
\end{framed}

{\bfseries\large Modifications}
\begin{framed}
  The CQ version of $q9$ is obtained by removing the
  {\bfseries ORDER BY} clause.
\end{framed}

\end{tcolorbox}
\egroup 
\caption{NPD Query q9}
\label{lst:npd-q9}
\end{lstfloat}

\pagebreak

\begin{lstfloat}[H]
\bgroup 
\singlespacing
\begin{tcolorbox}

{\bfseries\large NPD query q10}
\begin{framed}
\begin{lstlisting}[language=SPARQL,basicstyle=\ttfamily\scriptsize]
SELECT DISTINCT *
WHERE {
  [] a npdv:DiscoveryWellbore ;
  npdv:name ?wellbore; 
  npdv:dateUpdated ?date .
  FILTER (?date > "2013-01-01T00:00:00.0"^^xsd:dateTime)
} ORDER BY ?wellbore

\end{lstlisting}
\end{framed}

{\bfseries\large CQ query q10}
\begin{framed}
\begin{lstlisting}[language=SPARQL,basicstyle=\ttfamily\scriptsize]
SELECT DISTINCT *
WHERE {
  [] a npdv:DiscoveryWellbore ;
  npdv:name ?wellbore; 
  npdv:dateUpdated ?date .
  FILTER (?date > "2013-01-01T00:00:00.0"^^xsd:dateTime)
}

\end{lstlisting}
\end{framed}

{\bfseries\large Modifications}
\begin{framed}
  The CQ version of $q10$ is obtained by removing the
  {\bfseries ORDER BY} clause.
\end{framed}

\end{tcolorbox}
\egroup 
\caption{NPD Query q10}
\label{lst:npd-q10}
\end{lstfloat}

\pagebreak

\begin{lstfloat}[H]
\bgroup 
\singlespacing
\begin{tcolorbox}

{\bfseries\large NPD query q11}
\begin{framed}
\begin{lstlisting}[language=SPARQL,basicstyle=\ttfamily\scriptsize]
SELECT DISTINCT ?wellbore (?length AS ?lenghtM) ?company ?year 
WHERE {
  ?wc npdv:coreForWellbore
  [ rdf:type                      npdv:Wellbore ;
    npdv:name                     ?wellbore ;
    npdv:wellboreCompletionYear   ?year ;
    npdv:drillingOperatorCompany  [ npdv:name ?company ] 
  ] .
  { ?wc npdv:coresTotalLength ?length ;
    npdv:coreIntervalUOM "[m   ]"^^xsd:string .
  } 
  FILTER(?year >= 2008 &&
         ?length > 50)
} ORDER BY ?wellbore
\end{lstlisting}
\end{framed}

{\bfseries\large Description}
\begin{framed}
  Query $q11$ is a variation of $q6$ where the
  wellbore core length has a value expressed in meters.
\end{framed}
\end{tcolorbox}
\egroup 
\caption{NPD Query q11}
\label{lst:npd-q11}
\end{lstfloat}

\pagebreak

\begin{lstfloat}[H]
\bgroup 
\singlespacing
\begin{tcolorbox}

{\bfseries\large NPD query q12}
\begin{framed}
\begin{lstlisting}[language=SPARQL,basicstyle=\ttfamily\scriptsize]
SELECT DISTINCT ?wellbore (?length AS ?lenghtM) ?company ?year 
WHERE {
  ?wc npdv:coreForWellbore
  [ rdf:type                      npdv:Wellbore ;
    npdv:name                     ?wellbore ;
    npdv:wellboreCompletionYear   ?year ;
    npdv:drillingOperatorCompany  [ npdv:name ?company ] 
  ] .
  { ?wc npdv:coresTotalLength ?l ;
    npdv:coreIntervalUOM "[m   ]"^^xsd:string .
    BIND(?l AS ?length)
  } 

  UNION
  { ?wc npdv:coresTotalLength ?l ;
    npdv:coreIntervalUOM "[ft   ]"^^xsd:string .
    BIND((?l * 0.3048) AS ?length) 
  }
  FILTER(?year >= "2008"^^xsd:integer &&
         ?length > 50)
} ORDER BY ?wellbore
\end{lstlisting}
\end{framed}

{\bfseries\large Description}
\begin{framed}
  Query $q12$ is an extension of $q6$, and makes use of the SPARQL operator
  {\bfseries BIND} and arithmetical operations on the results,
  which are not supported on Mastro.
\end{framed}

\end{tcolorbox}
\egroup 
\caption{NPD Query q12}
\label{lst:npd-q12}
\end{lstfloat}

\pagebreak

\begin{lstfloat}[H]
\bgroup 
\singlespacing
\begin{tcolorbox}

{\bfseries\large NPD query q13}
\begin{framed}
\begin{lstlisting}[language=SPARQL,basicstyle=\ttfamily\scriptsize]
SELECT DISTINCT *
WHERE {
  ?x a npdv:SeismicSurvey .
  OPTIONAL {?x npdv:lengthCdpTotalKm ?cdpKM .}
  OPTIONAL {?x npdv:lengthBoatTotalKm ?boatKM .}
  FILTER (?cdpKM > 3660)
}

\end{lstlisting}
\end{framed}

{\bfseries\large NPD query q14}
\begin{framed}
\begin{lstlisting}[language=SPARQL,basicstyle=\ttfamily\scriptsize]
SELECT DISTINCT *
WHERE {
  ?x a npdv:WellboreDrillingMudSample ;
  npdv:dateMudMeasured ?date .
  OPTIONAL {
	?x npdv:mudType ?type .
	OPTIONAL {
	  ?x npdv:mudWeight ?w ;
	  npdv:mudMeasuredDepth ?d .
	} 
  }
  FILTER (?date > "1986-08-25T00:00:00"^^xsd:dateTime)
}
\end{lstlisting}
\end{framed}

{\bfseries\large Description}
\begin{framed}
  Queries $q_{13}$ and $q_{14}$ make use of the the
  SPARQL operators {\bfseries OPTIONAL} and {\bfseries BIND}
  which are not supported in the current version of Mastro.
\end{framed}
\end{tcolorbox}
\egroup 
\caption{NPD Query q13-14}
\label{lst:npd-q13-14}
\end{lstfloat}

\pagebreak

\begin{lstfloat}[H]
\bgroup 
\singlespacing
\begin{tcolorbox}

{\bfseries\large NPD query q15}
\begin{framed}
\begin{lstlisting}[language=SPARQL,basicstyle=\ttfamily\scriptsize]
SELECT  ?licenceURI (AVG(?interest) AS ?vavg)    
WHERE {   
  ?licenceURI a npdv:ProductionLicence .   
  [ ] a npdv:ProductionLicenceLicensee ;   
  npdv:dateLicenseeValidFrom ?date ;   
  npdv:licenseeInterest ?interest ;   
  npdv:licenseeForLicence ?licenceURI . 
  FILTER(?date >= "1979-12-31"^^xsd:date)  
} GROUP BY ?licenceURI  
\end{lstlisting}
\end{framed}

{\bfseries\large CQ query q15}
\begin{framed}
\begin{lstlisting}[language=SPARQL,basicstyle=\ttfamily\scriptsize]
SELECT  ?licenceURI ?interest
WHERE {   
  ?licenceURI a npdv:ProductionLicence .   
  [ ] a npdv:ProductionLicenceLicensee ;   
  npdv:dateLicenseeValidFrom ?date ;   
  npdv:licenseeInterest ?interest ;   
  npdv:licenseeForLicence ?licenceURI . 
  FILTER(?date >= "1979-12-31"^^xsd:date)  
}
\end{lstlisting}
\end{framed}

{\bfseries\large Modifications}
\begin{framed}
  The CQ of query $q15$ is obtained by removing the aggregation
  over the \verb|?licenceURI| variable.
\end{framed}

{\bfseries\large Description}
\begin{framed}
  Query $q15$ asks for the licenses, the interest applied to
  the respective licensees.
\end{framed}
\end{tcolorbox}
\egroup 
\caption{NPD Query q15}
\label{lst:npd-q15}
\end{lstfloat}

\pagebreak

\begin{lstfloat}[H]
\bgroup 
\singlespacing
\begin{tcolorbox}

{\bfseries\large NPD query q16}
\begin{framed}
\begin{lstlisting}[language=SPARQL,basicstyle=\ttfamily\scriptsize]
SELECT (COUNT(?licence ) AS ?licnumber)  
WHERE {   
  [ ] a npdv:ProductionLicence ;  
  npdv:name ?licence ;  
  npdv:dateLicenceGranted ?dateGranted ;  
  FILTER(?dateGranted >= "1999-12-31"^^xsd:date)  
} 
\end{lstlisting}
\end{framed}

{\bfseries\large CQ query q16}
\begin{framed}
\begin{lstlisting}[language=SPARQL,basicstyle=\ttfamily\scriptsize]
SELECT ?licence  
WHERE {   
  [ ] a npdv:ProductionLicence ;  
  npdv:name ?licence ;  
  npdv:dateLicenceGranted ?dateGranted ;  
  FILTER(?dateGranted >= "1999-12-31"^^xsd:date)  
} 
\end{lstlisting}
\end{framed}

{\bfseries\large Modifications}
\begin{framed}
The CQ of query $q16$ is obtained by removing
the {\bfseries COUNT} operator.
\end{framed}

{\bfseries\large Description}
\begin{framed}
 This query asks for the name of the licenses granted starting from year $2000$.
\end{framed}
\end{tcolorbox}
\egroup 
\caption{NPD Query q16}
\label{lst:npd-q16}
\end{lstfloat}

\pagebreak

\begin{lstfloat}[H]
\bgroup 
\singlespacing
\begin{tcolorbox}

{\bfseries\large NPD query q17}
\begin{framed}
\begin{lstlisting}[language=SPARQL,basicstyle=\ttfamily\scriptsize]
SELECT ?field (SUM(?g) AS ?gas)
WHERE {
  [ npdv:productionYear 2008 ;
    npdv:productionMonth ?m ;
    npdv:producedGas     ?g ;
    npdv:productionForField
    [ rdf:type npdv:Field ;
      npdv:name ?field ;
      npdv:currentFieldOperator
      [ npdv:shortName "STATOIL PETROLEUM AS"^^xsd:string ]]]
} GROUP BY ?field ORDER BY ?field
\end{lstlisting}
\end{framed}

{\bfseries\large CQ query q17}
\begin{framed}
\begin{lstlisting}[language=SPARQL,basicstyle=\ttfamily\scriptsize]
SELECT ?field ?g
WHERE {
  [ npdv:productionYear 2008 ;
    npdv:productionMonth ?m ;
    npdv:producedGas     ?g ;
    npdv:productionForField
    [ rdf:type npdv:Field ;
      npdv:name ?field ;
      npdv:currentFieldOperator
      [ npdv:shortName "STATOIL PETROLEUM AS"^^xsd:string ]]]
}
\end{lstlisting}
\end{framed}

{\bfseries\large Modifications}
\begin{framed}
  Query $q17$ is obtained by removing aggregation
  and the ordering over the \verb|?field| variable.
\end{framed}

\end{tcolorbox}
\egroup 
\caption{NPD Query q17}
\label{lst:npd-q17}
\end{lstfloat}

\pagebreak

\begin{lstfloat}[H]
\bgroup 
\singlespacing
\begin{tcolorbox}

{\bfseries\large NPD query q18}
\begin{framed}
\begin{lstlisting}[language=SPARQL,basicstyle=\ttfamily\scriptsize]
SELECT ?field (AVG(?oil) AS ?avgOil)  
WHERE {
  [ ] a npdv:FieldYearlyProduction ;
  npdv:productionForField [ npdv:name ?field ] ;
  npdv:producedOil            ?oil ;
  npdv:productionYear         ?year .
  FILTER (?year < 2013)
} 
GROUP BY ?field
ORDER BY DESC(?avgOil)
\end{lstlisting}
\end{framed}

{\bfseries\large CQ query q18}
\begin{framed}
\begin{lstlisting}[language=SPARQL,basicstyle=\ttfamily\scriptsize]
SELECT ?field ?oil
WHERE {
   [ ] a npdv:FieldYearlyProduction ;
       npdv:productionForField [ npdv:name ?field ] ;
       npdv:producedOil            ?oil ;
       npdv:productionYear         ?year .
   FILTER (?year < 2013)
} 
\end{lstlisting}
\end{framed}

{\bfseries\large Modifications}
\begin{framed}
  The CQ of query $q18$ is obtained by removing the aggregation
  over the \verb|?field| variable and the ordering over the averaged
  \verb|?oil| variable.
\end{framed}

\end{tcolorbox}
\egroup 
\caption{NPD Query q18}
\label{lst:npd-q18}
\end{lstfloat}

\pagebreak

\begin{lstfloat}[H]
\bgroup 
\singlespacing
\begin{tcolorbox}

{\bfseries\large NPD query q19}
\begin{framed}
\begin{lstlisting}[language=SPARQL,basicstyle=\ttfamily\scriptsize]
SELECT ?field (SUM(?o) AS ?oil)
WHERE {
  [ npdv:productionYear 1993 ;
    npdv:productionMonth ?m ;
    npdv:producedOil     ?o ;
    npdv:productionForField
    [ rdf:type npdv:Field ;
      npdv:name ?field ;
      npdv:currentFieldOperator
      [ npdv:shortName "STATOIL PETROLEUM AS"^^xsd:string ]]]
  FILTER(?m >= 1 &&
         ?m <= 6)
} GROUP BY ?field ORDER BY ?field
\end{lstlisting}
\end{framed}

{\bfseries\large CQ query q19}
\begin{framed}
\begin{lstlisting}[language=SPARQL,basicstyle=\ttfamily\scriptsize]
SELECT ?field ?o
WHERE {
  [ npdv:productionYear 1993 ;
    npdv:productionMonth ?m ;
    npdv:producedOil     ?o ;
    npdv:productionForField
    [ rdf:type npdv:Field ;
      npdv:name ?field ;
      npdv:currentFieldOperator
      [ npdv:shortName "STATOIL PETROLEUM AS"^^xsd:string ]]]
  FILTER(?m >= 1 &&
         ?m <= 6)
}
\end{lstlisting}
\end{framed}

{\bfseries\large Modifications}
\begin{framed}
  The CQ of query $q19$ is obtained by removing the aggregation
  and ordering over the \verb|?field| variable.
\end{framed}

\end{tcolorbox}
\egroup 
\caption{NPD Query q19}
\label{lst:npd-q19}
\end{lstfloat}

\pagebreak

\begin{lstfloat}[H]
\bgroup 
\singlespacing
\begin{tcolorbox}

{\bfseries\large NPD query q20}
\begin{framed}
\begin{lstlisting}[language=SPARQL,basicstyle=\ttfamily\scriptsize]
SELECT ?fr (Max(?g) AS ?max)   
WHERE {   
  ?fr npdv:productionYear ?year ;   
  npdv:productionMonth ?m ;   
  npdv:producedGas     ?g .
  FILTER (?year > 2000)      
} GROUP BY ?fr
\end{lstlisting}
\end{framed}

{\bfseries\large NPD query q21}
\begin{framed}
\begin{lstlisting}[language=SPARQL,basicstyle=\ttfamily\scriptsize]
SELECT ?fr (Min(?g) AS ?min)   
WHERE {   
  ?fr npdv:productionYear ?year ;   
  npdv:productionMonth ?m ;   
  npdv:producedGas     ?g .
  FILTER (?year > 2000)      
} GROUP BY ?fr
\end{lstlisting}
\end{framed}

{\bfseries\large CQ query q20}
\begin{framed}
\begin{lstlisting}[language=SPARQL,basicstyle=\ttfamily\scriptsize]
SELECT ?fr ?g   
WHERE {   
  ?fr npdv:productionYear ?year ;   
  npdv:productionMonth ?m ;   
  npdv:producedGas     ?g .
  FILTER (?year > 2000)      
}
\end{lstlisting}
\end{framed}

{\bfseries\large Modifications}
\begin{framed}
Query $q20$ and $q21$ differ only by the type of aggregation performed.
For this reason after removing the aggregation only one of them has been considered.
\end{framed}

\end{tcolorbox}
\egroup 
\caption{NPD Query q20}
\label{lst:npd-q20}
\end{lstfloat}

\pagebreak

\begin{lstfloat}[H]
\bgroup 
\singlespacing
\begin{tcolorbox}

{\bfseries\large NPD query q22}
\begin{framed}
\begin{lstlisting}[language=SPARQL,basicstyle=\ttfamily\scriptsize]
SELECT DISTINCT ?wc 
WHERE { 
  ?wc npdv:coreForWellbore [ rdf:type npdv:Wellbore ]. 
}
\end{lstlisting}
\end{framed}

{\bfseries\large NPD query q23}
\begin{framed}
\begin{lstlisting}[language=SPARQL,basicstyle=\ttfamily\scriptsize]
SELECT DISTINCT ?member ?awc   
WHERE {
  ?member npdv:member ?awc.
  ?awc npdv:coreForWellbore [ rdf:type npdv:Wellbore ].   
}
\end{lstlisting}
\end{framed}

{\bfseries\large NPD query q24}
\begin{framed}
\begin{lstlisting}[language=SPARQL,basicstyle=\ttfamily\scriptsize]
SELECT DISTINCT ?member   
WHERE {
  ?member npdv:member ?awc.
  ?awc npdv:coreForWellbore [ rdf:type npdv:Wellbore ].   
}
\end{lstlisting}
\end{framed}

{\bfseries\large NPD query q25}
\begin{framed}
\begin{lstlisting}[language=SPARQL,basicstyle=\ttfamily\scriptsize]
SELECT DISTINCT ?licensee   
WHERE {
  ?licensee npdv:licenseeForLicence ?licence.
}
\end{lstlisting}
\end{framed}

{\bfseries\large Description}
\begin{framed}
  Queries $q22-25$ are already in the fragment of conjuntive queries,
  so no modification is required.
\end{framed}
\end{tcolorbox}
\egroup 
\caption{NPD Query q22-25}
\label{lst:npd-q22-25}
\end{lstfloat}

\pagebreak

\begin{lstfloat}[H]
\bgroup 
\singlespacing
\begin{tcolorbox}

{\bfseries\large NPD query q26}
\begin{framed}
\begin{lstlisting}[language=SPARQL,basicstyle=\ttfamily\scriptsize]
SELECT DISTINCT ?licensee
WHERE {
  ?licensee npdv:licenseeForLicence
    [ npdv:licenceOperatorCompany ?company ]
}
\end{lstlisting}
\end{framed}

{\bfseries\large NPD query q27}
\begin{framed}
\begin{lstlisting}[language=SPARQL,basicstyle=\ttfamily\scriptsize]
SELECT DISTINCT ?company
WHERE {
  ?licensee npdv:licenseeForLicence
    [ npdv:licenceOperatorCompany ?company ].
}
\end{lstlisting}
\end{framed}

{\bfseries\large Description}
\begin{framed}
  Queries $q26-27$ are already in the fragment of conjuntive queries,
  so no modification is required.
\end{framed}
\end{tcolorbox}
\egroup 
\caption{NPD Query q26-27}
\label{lst:npd-q26-27}
\end{lstfloat}

\pagebreak

\begin{lstfloat}[H]
\bgroup 
\singlespacing
\begin{tcolorbox}

{\bfseries\large NPD query q28}
\begin{framed}
\begin{lstlisting}[language=SPARQL,basicstyle=\ttfamily\scriptsize]
SELECT DISTINCT ?wellbore ?wc ?well 
WHERE { 
  ?wellbore npdv:wellboreForDiscovery ?discovery;
		    npdv:belongsToWell ?well.  
  ?wc npdv:coreForWellbore ?wellbore.
}
\end{lstlisting}
\end{framed}

{\bfseries\large NPD query q29}
\begin{framed}
\begin{lstlisting}[language=SPARQL,basicstyle=\ttfamily\scriptsize]
SELECT DISTINCT ?wellbore ?wc ?well ?length
WHERE { 
  ?wellbore npdv:wellboreForDiscovery ?discovery;
		    npdv:belongsToWell ?well.  
  ?wc npdv:coreForWellbore ?wellbore;
	  npdv:coresTotalLength ?length;
   npdv:coreIntervalUOM "[ft   ]"^^xsd:string . # feets
  
  FILTER (?length < 56796)		   
}
\end{lstlisting}
\end{framed}

{\bfseries\large Description}
\begin{framed}
  Queries $q28-29$ are already in the fragment of conjuntive queries,
  so no modification is required.
\end{framed}
\end{tcolorbox}
\egroup 
\caption{NPD Query q28-29}
\label{lst:npd-q28-29}
\end{lstfloat}

\pagebreak

\begin{lstfloat}[H]
\bgroup 
\singlespacing
\begin{tcolorbox}

{\bfseries\large NPD query q30}
\begin{framed}
\begin{lstlisting}[language=SPARQL,basicstyle=\ttfamily\scriptsize]
SELECT DISTINCT ?wellbore ?wc ?well ?length
WHERE { 
  ?wellbore npdv:wellboreForDiscovery ?discovery;
            npdv:belongsToWell ?well.  
  ?wc npdv:coreForWellbore ?wellbore.
  {  
    ?wc npdv:coresTotalLength ?lmeters ;
    npdv:coreIntervalUOM "[m   ]"^^xsd:string .
    BIND(?lmeters AS ?length)
  } 
  UNION
  {
    ?wc npdv:coresTotalLength ?lfeets ;
    npdv:coreIntervalUOM "[ft   ]"^^xsd:string .
    BIND((?lfeets * 0.3048) AS ?length) 
  }                
  FILTER (?length < 22337)
}
\end{lstlisting}
\end{framed}

{\bfseries\large Description}
\begin{framed}
  Query $q30$ is an extension of $q12$, and makes use of the SPARQL operator
  {\bfseries BIND} and arithmetical operations on the results,
  which are not supported on Mastro.
\end{framed}
\end{tcolorbox}
\egroup 
\caption{NPD Query q30}
\label{lst:npd-q30}
\end{lstfloat}

\pagebreak

\paragraph{Data generation}
\label{sec:npd-datagen}
The original NPD databases is derived from the data published on the
{\em Norvegian Petroleum Directorate}
FactPages\footnote{http://factpages.npd.no/factpages/}.

The data from FactPages has been translated from CSV files
into a structured database. The generated schema consists of $70$ tables
with $276$ distinct columns (about $1000$ columns in total),
and $94$ foreign keys.

The schemas of the tables overlap in the sense that several attributes
are replicated in several tables. In fact, there are tables with more
than $100$ columns. The total size of the initial database is about $60$ MB.

Since OBDA are expected to work in the context of Big Data, the
authors of the benchmark have provided a tool that enables the initial
database instance to be scaled in order to obtain larger instances.
The scaling process, implemented by the {\em Virtual Instances Generator}
(VIG)\cite{LantiXC16}\footnote{https://github.com/ontop/vig},
is performed by taking into account the axioms in the ontology,
the structure of the mappings, and the database constraints in order to preserve
a set of similarity measures in the original database.

Compared to a random generation approach, the algorithm adopted
by VIG preserves important factors, such as the ratio of
column-based duplicates, null values ratios,
and the ratio for join result sets. This is a requirement
when the generation has to be performed in order to evaluate the
performance of an OBDA system. Adopting a completely random
approach would simply make the scaled database non-suitable
for the evaluation of an OBDA system, since the number of joining columns
in the mappings would be completely random and non-representative
of the original instance.

Starting from this initial database, instances of different size
have been created with the use of the VIG generator, and have
been loaded into separate databases.
Table~\ref{tab:generated-databases} shows the scaling factor and the
size for each of the generated databases that have been
used in our experiment. The number of the database represents its
scale with respect to the original instance.

\vspace{2em}

\begin{table}[hbpt]
  \centering
  \input{npd-databases}
  \caption{Generated databases}
  \label{tab:generated-databases}
\end{table}

\vfill
\pagebreak

\paragraph{Experimental Setup}
\label{sec:npd-results}
We ran the NPD Benchmark on both the latest version of Mastro $1.0.2$
and Ontop $3.0$, on the same physical system using the same set of
generated databases. The underlying DBMS is MySQL version $5.7.21$,
running locally on the testing machine.

The specifications of the platform used for the experiments are the following:
\begin{description}
\item[CPU] Intel Xeon E5-2670 running at 2.60GHz
\item[RAM] 16 GB DDR3 1600 MHz
\item[OS] Ubuntu 17.10 running in a virtualized environment (4 cores)
\end{description}

The experimentation is performed in the following mode:
\begin{itemize}
\item We iterate over the set of queries. At each iteration we pick
  a random query from the set and evaluate it over the system.
  This is done in order to reduce the effect of the caching in the DBMS.

  For each execution we store the results and the time it took
  to complete. We consider the combined time needed for evaluating
  the query and processing the set of results.
  Queries are executed sequentially through the
  use of a testing platform designed specifically for this
  task, which accesses directly the internal APIs of the systems.
\item We repeat the process until all queries have been executed
  a fixed amount of time, in this case $5$ executions were performed.
\item Finally, we take the average of the execution times for
  each of the queries in the set.
\end{itemize}

\begin{figure}[H]
  \centering
  \includegraphics[width=0.9\textwidth]{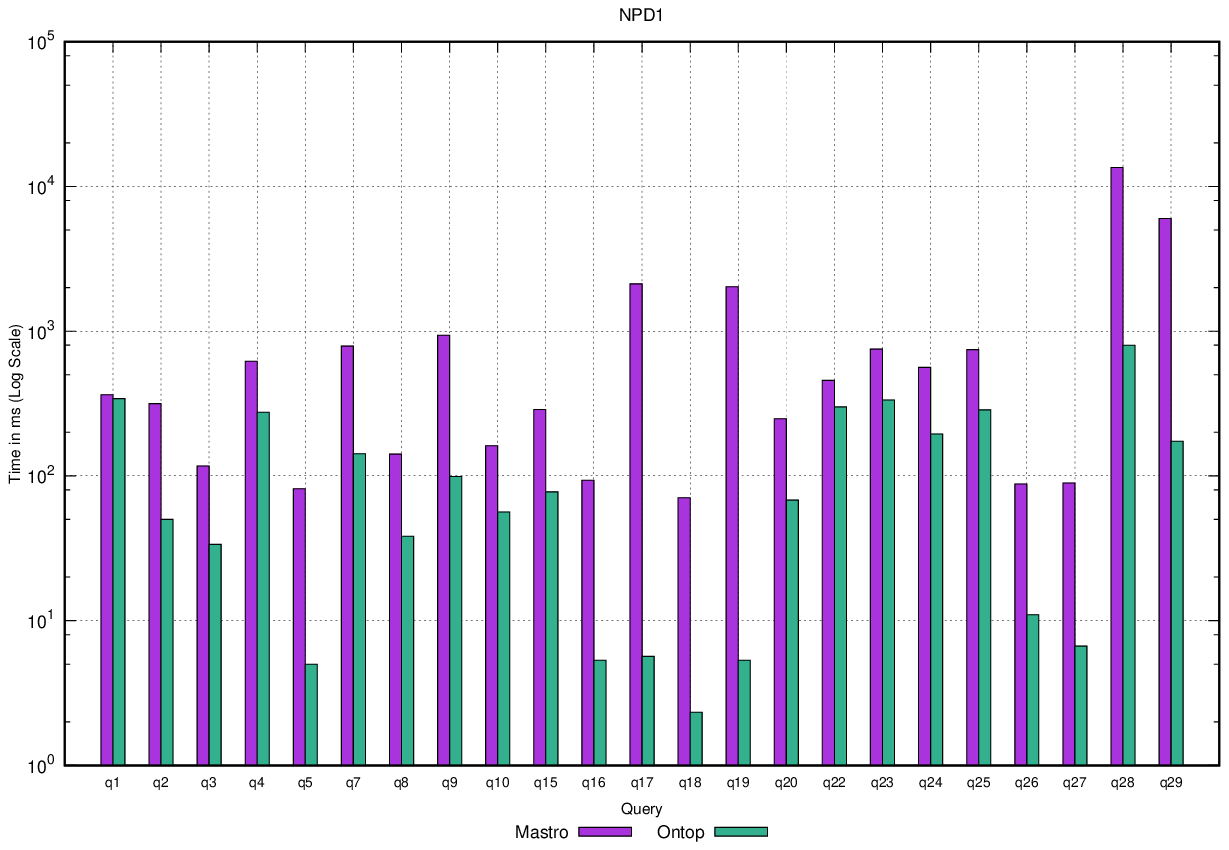}
  \caption{Query Answering over NPD1 (Execution Time)}
  \label{fig:total-time-npd1}
\end{figure}

\begin{figure}[H]
  \centering
  \includegraphics[width=0.9\textwidth]{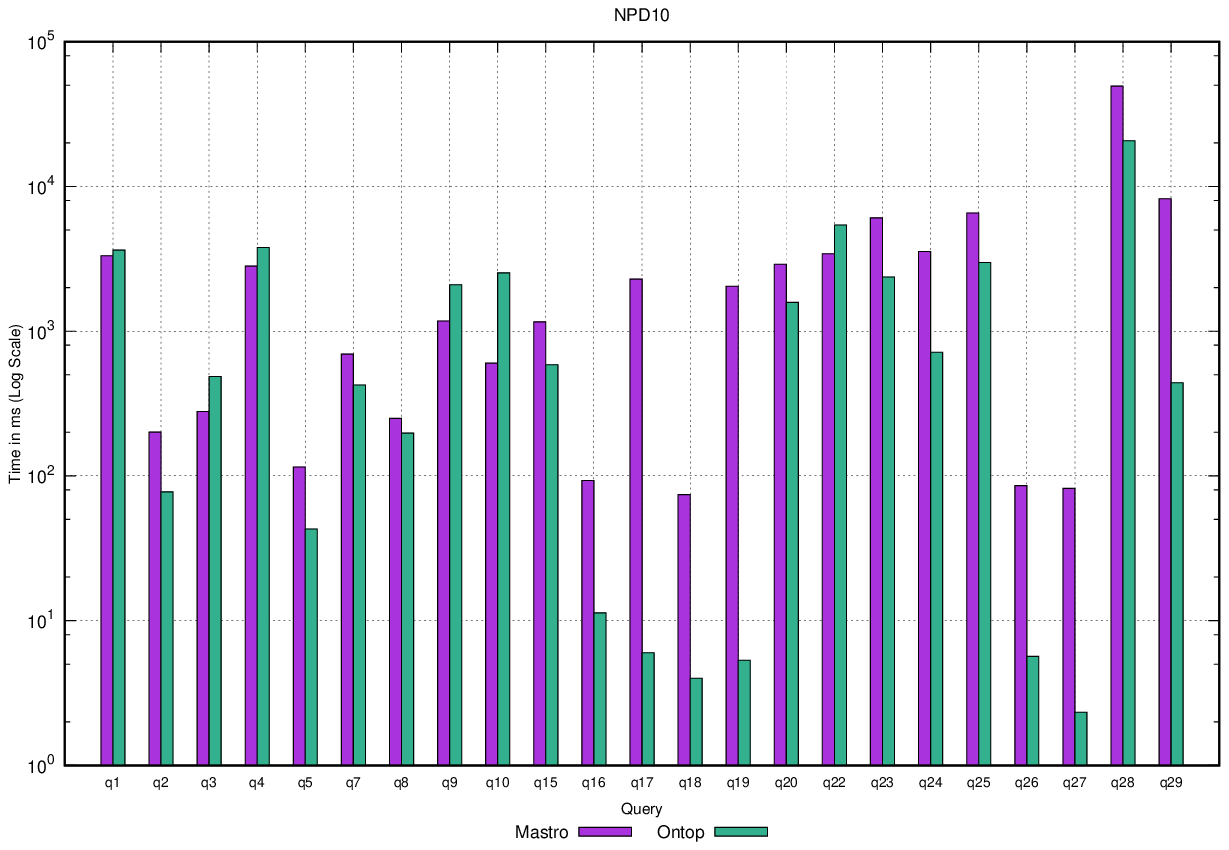}
  \caption{Query Answering over NPD10 (Execution Time)}
  \label{fig:total-time-npd10}
\end{figure}

\begin{figure}[H]
  \centering
  \includegraphics[width=0.9\textwidth]{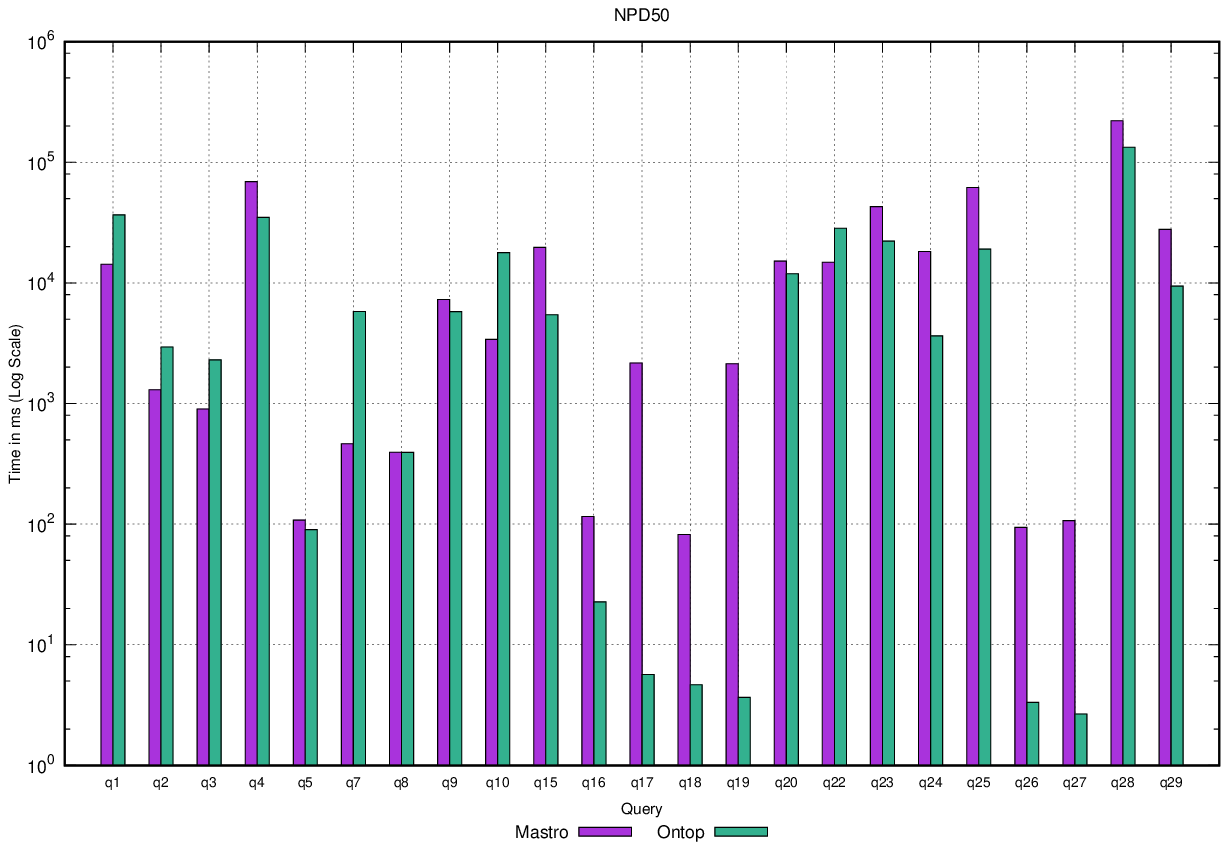}
  \caption{Query Answering over NPD50 (Execution Time)}
  \label{fig:total-time-npd50}
\end{figure}

\begin{figure}[H]
  \centering
  \includegraphics[width=0.9\textwidth]{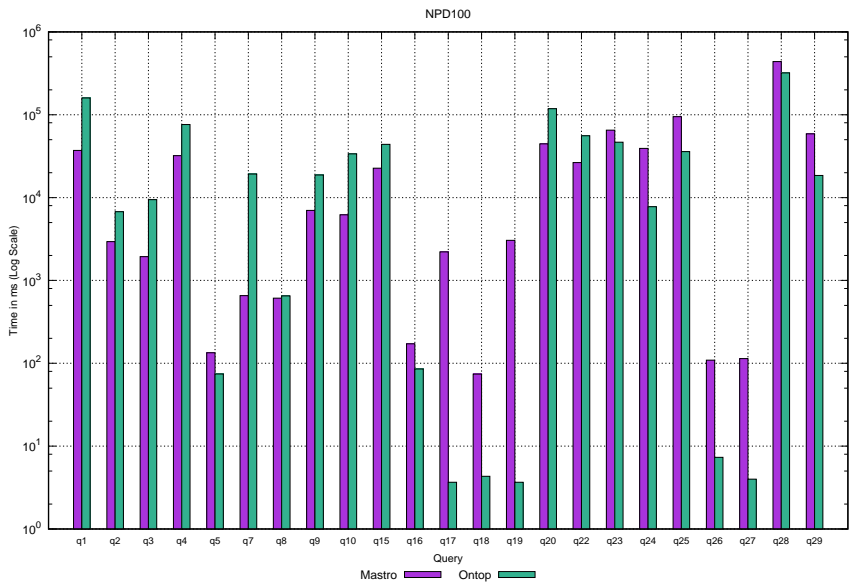}
  \caption{Query Answering over NPD100 (Execution Time)}
  \label{fig:total-time-npd100}
\end{figure}

In order to compare the systems under the same setting
we enabled for both reasoning with respect to existential variables.
Our metric of comparison is the total time taken to complete
the execution of the query and to process the results.
A comparison of the execution times for both systems
for the database instances NPD1, NPD10, NPD50, and NPD100
are shown respectively in Figure~\ref{fig:total-time-npd1},
\ref{fig:total-time-npd10}, \ref{fig:total-time-npd50},
and~\ref{fig:total-time-npd100}.
Table~\ref{tab:total-time-comparison} reports a summary of the average
execution time for each query and database size.

\vfill

\begin{sidewaystable}[hbpt]
  \centering
  \caption{Query Answering Times over the generated databases (in seconds)}
  \label{tab:total-time-comparison}
  \input{npd-times}

\end{sidewaystable}

\pagebreak

\paragraph{Experimental results}
We start our analysis by looking at some of the most interesting results.
In particular, the queries where Ontop performs worse are those requiring
reasoning with respect to the existential variables.
Examples of these queries are $q9$ and $q10$, which cause the system
to produce rewritings consisting of unions of tens of sub-queries.

Queries $q1$,$q2$,$q3$,$q4$,$q7$,$q15$,$q16$,$q25$ instead produce a simple SPJ
rewriting, although the difference here is given by the fact that Ontop
applies strictly the rules specified by the OWL 2 standard,
performing the datatype casts and the generation of the IRIs
directly at the SQL level, which causes a slowdown of the execution.
Instead, Mastro adopts a less strict approach for dealing with
datatypes, avoiding to perform casts and IRI-construction at the SQL level.

As for queries $q24$-$q25$, and $q28$-$q29$, it can be noted that
in this case Mastro performs considerably slower than Ontop.
This is due to the high number of mappings for the ontology predicates
involved in these queries, that have to be unfolded by the system,
and by the high number of sub-classes and sub-properties
for the concepts and properties in the query,
which cause the rewriting to grow exponentially in size.
In Mastro, this phenomenon is mitigated with the addition of data constraints
during the design of the views, in such a way that this redundancy is avoided.
In this case, since view are generated automatically from the R2RML mapping,
these constraints are not available to the system, effectively disabling
all the optimizations that the system is capable of performing.
Instead, in Ontop this problem is mitigated during the offline stage,
when the mappings and the ontology
are compiled to form the so-called $\mathcal{T}$-mappings, using
the database constraints to optimize the mappings.

Finally, queries $q17$-$q19$, and $q26$-$q27$ are less interesting, as they
produce empty unfoldings (due to mismatching function names in the mappings),
and their execution time does not depend on the database size.
For these queries, it can be noted that moving part of the
computation to an offline stage gives Ontop a big advantage, as
the system has to check a small amount of mappings at query execution time.


%% file: npd-databases.tex

\begin{tabular}{|c|c|c|}
  \hline
  \multicolumn{1}{|c|}{\textsc{\bfseries Name}} &
  \multicolumn{1}{|c|}{\textsc{\bfseries Scale Factor}} &
  \multicolumn{1}{|c|}{\textsc{\bfseries Size}}\\ \hline
  NPD1   & 1   & 60 MB   \\ \hline
  NPD10  & 10  & 710 MB  \\ \hline
  NPD50  & 50  & 2570 MB \\ \hline
  NPD100 & 100 & 5300 MB \\ \hline
\end{tabular}


%% file: npd-times.tex

\begin{tabular}{c|l|l|l|l|l|l|l|l}
  \hline
  \multicolumn{1}{|c|}{\textsc{\bfseries Query}} &
  \multicolumn{2}{|c|}{\textsc{\bfseries NPD1}} &
  \multicolumn{2}{|c|}{\textsc{\bfseries NPD10}} &
  \multicolumn{2}{|c|}{\textsc{\bfseries NPD50}} &
  \multicolumn{2}{|c|}{\textsc{\bfseries NPD100}}\\ \hline\hline
        & Mastro    & Ontop    & Mastro    & Ontop     & Mastro     & Ontop      & Mastro     & Ontop      \\ \hline \hline
  $q1$  & $0.364  $ & $0.692 $ & $3.320  $ & $3.627  $ & $14.277  $ & $36.657  $ & $37.107  $ & $159.733 $ \\ \hline
  $q2$  & $0.316  $ & $0.098 $ & $0.201  $ & $0.078  $ & $1.302   $ & $2.944   $ & $2.939   $ & $6.762   $ \\ \hline
  $q3$  & $0.117  $ & $0.064 $ & $0.279  $ & $0.485  $ & $0.900   $ & $2.304   $ & $1.939   $ & $9.450   $ \\ \hline
  $q4$  & $0.620  $ & $0.996 $ & $2.815  $ & $3.779  $ & $69.200  $ & $34.961  $ & $32.037  $ & $76.069  $ \\ \hline
  $q5$  & $0.082  $ & $0.024 $ & $0.115  $ & $0.043  $ & $0.108   $ & $0.090   $ & $0.134   $ & $0.074   $ \\ \hline
  $q7$  & $0.788  $ & $0.258 $ & $0.694  $ & $0.425  $ & $0.464   $ & $5.793   $ & $0.656   $ & $19.252  $ \\ \hline
  $q8$  & $0.141  $ & $0.076 $ & $0.250  $ & $0.197  $ & $0.395   $ & $0.394   $ & $0.612   $ & $0.651   $ \\ \hline
  $q9$  & $0.937  $ & $0.125 $ & $1.176  $ & $2.096  $ & $7.284   $ & $5.788   $ & $7.012   $ & $18.753  $ \\ \hline
  $q10$ & $0.161  $ & $0.250 $ & $0.600  $ & $2.523  $ & $3.401   $ & $17.849  $ & $6.207   $ & $33.731  $ \\ \hline
  $q15$ & $0.288  $ & $0.111 $ & $1.161  $ & $0.586  $ & $19.688  $ & $5.454   $ & $22.683  $ & $43.994  $ \\ \hline
  $q16$ & $0.093  $ & $0.011 $ & $0.093  $ & $0.011  $ & $0.116   $ & $0.023   $ & $0.172   $ & $0.085   $ \\ \hline
  $q17$ & $2.121  $ & $0.004 $ & $2.287  $ & $0.006  $ & $2.167   $ & $0.006   $ & $2.213   $ & $0.004   $ \\ \hline
  $q18$ & $0.071  $ & $0.009 $ & $0.074  $ & $0.004  $ & $0.082   $ & $0.005   $ & $0.074   $ & $0.004   $ \\ \hline
  $q19$ & $2.024  $ & $0.005 $ & $2.043  $ & $0.005  $ & $2.138   $ & $0.004   $ & $3.051   $ & $0.004   $ \\ \hline
  $q20$ & $0.249  $ & $0.113 $ & $2.901  $ & $1.579  $ & $15.146  $ & $11.898  $ & $44.538  $ & $118.329 $ \\ \hline
  $q22$ & $0.457  $ & $0.565 $ & $3.421  $ & $5.408  $ & $14.808  $ & $28.372  $ & $26.425  $ & $55.841  $ \\ \hline
  $q23$ & $0.752  $ & $0.243 $ & $6.065  $ & $2.363  $ & $42.884  $ & $22.180  $ & $65.265  $ & $46.621  $ \\ \hline
  $q24$ & $0.562  $ & $0.114 $ & $3.546  $ & $0.715  $ & $18.224  $ & $3.640   $ & $39.278  $ & $7.772   $ \\ \hline
  $q25$ & $0.745  $ & $0.284 $ & $6.551  $ & $2.985  $ & $61.744  $ & $19.048  $ & $94.802  $ & $35.893  $ \\ \hline
  $q26$ & $0.088  $ & $0.004 $ & $0.086  $ & $0.006  $ & $0.094   $ & $0.003   $ & $0.109   $ & $0.007   $ \\ \hline
  $q27$ & $0.089  $ & $0.003 $ & $0.082  $ & $0.002  $ & $0.107   $ & $0.003   $ & $0.114   $ & $0.004   $ \\ \hline
  $q28$ & $13.496 $ & $1.648 $ & $49.303 $ & $20.641 $ & $220.932 $ & $133.396 $ & $438.149 $ & $319.339 $ \\ \hline
  $q29$ & $6.001  $ & $0.155 $ & $8.211  $ & $0.440  $ & $27.770  $ & $9.396   $ & $58.841  $ & $18.525  $ \\ \hline
\end{tabular}

%% file: sec-aci-prep.tex

\section{Comparison on a full-fledged OBDA solution}
\label{sec:aci}
While the use synthetic benchmarks can be a starting point
in order to understand the type of performance that
we can expect from a system, and allows to draw some
conclusions about the effectiveness of the adopted approach,
there is also the need to verify that such conclusions
hold also when we move to real world applications.

The main problem with the specification that we presented in
Section~\ref{sec:npd-benchmark} is that while the ontology
has been effectively designed to model accurately the conceptual
reality of the domain of interest, the data sources did not
exist before and have been created specifically for the
purpose of evaluating the specification. Unfortunately,
this means that they do not reflect
what is the real structure that can be found when attempting
to experiment the ontology-based data access approach
in real world scenarios.

For this purpose we decided to use one of the specifications
that is currently in development between Sapienza University
of Rome and the Automobile Club d'Italia (ACI).

In this chapter we first give an overview of the specification,
by describing the ontology, the mappings and the data sources.
Then, we proceed in showing the experimentation that we performed,
aimed at understanding the usability of both Mastro
and Ontop in such scenario. Finally, we present the results
and draw the conclusions.

\paragraph{Ontology}
The ontology comprises about $500$ concepts, $200$ roles and $200$ attributes,
and it is divided into $11$ logically interconnected modules.
Among these modules some of the most important are represented by:
\begin{itemize}
\item The ontological module describing the concept of \emph{Vehicle}
  (Veicolo),
  characterized as a central object in the ontology, modeling
  also its relevant state (\emph{Stato}), that models the
  evolution of vehicles' property over time.
\item The ontological module describing the concept of \emph{Subject} (Soggetto),
  modeling the possible roles played by the subjects
  (physical people or organizations) with respect to the taxation
  concerning vehicles.
\item The ontological module describing the concept of \emph{Formality}
  (Formalità), and \emph{payment} (Pagamento).
\end{itemize}

The ontology has been defined following rigorously the best practice
for ontology design, based on the analysis and definition of the domain
of interest through a series of interviews with the domain experts,
so that the ontology would reflect exactly the reality of the
domain of interest, and not the structure of the data sources.
This characteristics of the ontology makes so that the specification
of the mappings is extremely complex, due to the large \emph{semantic gap}
between the reality of the domain of interest and the
semantics of the data sources, which are structured in such a way
that enables the applications that make use of them meet the efficiency
requirements that they need.

Moreover, the modeling of the ontology also describes the evolution
in time of its elements. For example, during its lifespan
a vehicle can change owner, license plate, and so on.
This is captured by the notion of state of a vehicle (Stato).
This is a recurring aspect in the queries as we will show in the
following sections.

\paragraph{Data source}
For this experiment the database is managed by an instance of
Oracle 12c. The database is accessed remotely through the use of a VPN.
The relevant portion of the data for our experiment
is distributed across $6$ schemas.
These schemas are composed of hundreds of tables, but for what regards
the portion of the domain that is of interest in this experiment
we concentrate on about $90$ relational tables,
ranging from information regarding the domain of PRA (Pubblico Registro Automobilistico),
to those regarding the taxation concerning the vehicles.
Some of these tables count from $200$ million tuples up to above
$1$ billion tuples in some cases, with a number of attributes ranging from
$30$ to $100$. The overall size of the portion of interest
of the data source is several gigabytes of data (an accurate estimate
was not possible).

\paragraph{Mappings}
The specification comprises $976$ ontology mappings, composed from
a set of about $110$ views over the data source. About $300$
of these ontology mappings are built from single view atoms, while the remaining
are specified as conjunctive queries over the view predicates.
For the purpose of our experimentation the mappings have been
previously translated in R2RML through the use of the approach
described in Section~\ref{sec:r2rml}, and then imported back into both systems.
During the translation process the constraints over the view predicates
are discarded, since they cannot be expressed in R2RML.

\paragraph{Query Set}
In this section we describe the set of queries that have been defined
to evaluate the usability of both systems. Query $q1$ to $q5$ are basic
navigational queries, that span some of the relevant part of the ontology.
The remaining queries $q6$ to $q10$ are derived the real queries used
in the original experimentation of the project. These queries are built from
a set of \emph{competency questions}, defined by interviewing the
domain experts over the practical questions that have to be posed
to the system, and are used to validate the quality and the coherency
of the specification.

Before showing the set of queries that have been used for our experimentation
we have to point out that in order to comply with the time restrictions
we were granted, these queries have been restricted to extracting
information for single vehicles. In some cases this was already
enough to fill the time slot that was allowed.
In their real application, queries are used to build reports
for several hundred of thousands of vehicles, that require several
hours to complete.

\vfill
\pagebreak


\begin{lstfloat}[H]
\bgroup 
\singlespacing
\begin{tcolorbox}

{\bfseries\large ACI query q1}
\begin{framed}
\begin{lstlisting}[language=SPARQL,basicstyle=\ttfamily\scriptsize]
PREFIX aci: <http://www.aci.it/ontology#>
PREFIX xsd: <http://www.w3.org/2001/XMLSchema#>

SELECT ?veicolo ?stato ?proprieta
WHERE {
  ?veicolo aci:ID_veicolo "<id>"^^xsd:string.
  ?veicolo aci:ha_stato_di_veicolo ?stato.
  ?proprieta aci:proprieta_di ?veicolo.
}
\end{lstlisting}
\end{framed}

{\bfseries\large Description}
\begin{framed}
  Query $q_{1}$ asks for vehicles that have a state and their ownership.
\end{framed}
\end{tcolorbox}
\egroup 
\caption{ACI query q1}
\label{lst:aci-query-q1}
\end{lstfloat}


\begin{lstfloat}[H]
\bgroup 
\singlespacing
\begin{tcolorbox}

{\bfseries\large ACI query q2}
\begin{framed}
\begin{lstlisting}[language=SPARQL,basicstyle=\ttfamily\scriptsize]
PREFIX aci: <http://www.aci.it/ontology#>
PREFIX xsd: <http://www.w3.org/2001/XMLSchema#>

SELECT ?veicolo ?stato ?proprieta ?formalita
WHERE {
  ?veicolo aci:ID_veicolo "<id>"^^xsd:string.
  ?veicolo aci:ha_stato_di_veicolo ?stato.
  ?proprieta aci:proprieta_di ?veicolo.
  ?formalita aci:formalita_presentata_per_veicolo ?veicolo.
}
\end{lstlisting}
\end{framed}

{\bfseries\large Description}
\begin{framed}
  Query $q2$ asks for vehicles that have a state, their ownership, and
  the formalities.
\end{framed}
\end{tcolorbox}
\egroup 
\caption{ACI query q2}
\label{lst:aci-query-q2}
\end{lstfloat}


\begin{lstfloat}[H]
\bgroup 
\singlespacing
\begin{tcolorbox}

{\bfseries\large ACI query q3}
\begin{framed}
\begin{lstlisting}[language=SPARQL,basicstyle=\ttfamily\scriptsize]
PREFIX aci: <http://www.aci.it/ontology#>
PREFIX xsd: <http://www.w3.org/2001/XMLSchema#>

SELECT ?veicolo ?stato ?proprieta
       ?accettazione ?anno ?descrizione
WHERE {
  ?veicolo aci:ID_veicolo "<id>"^^xsd:string.
  ?veicolo aci:ha_stato_di_veicolo ?stato.
  ?proprieta aci:proprieta_di ?veicolo.
  ?formalita aci:formalita_presentata_per_veicolo ?veicolo.
  ?formalita aci:data_accettazione_formalita ?accettazione.
  ?formalita aci:anno_formalita ?anno.
  ?formalita aci:descrizione_tipo ?descrizione.
}
\end{lstlisting}
\end{framed}

{\bfseries\large Description}
\begin{framed}
  Query $q2$ asks for vehicles that have a state, their ownership, and
  the formalities, with their acceptance dates, year of acceptance,
  and a description of the type of formality.
\end{framed}
\end{tcolorbox}
\egroup 
\caption{ACI query q3}
\label{lst:aci-query-q3}
\end{lstfloat}


\begin{lstfloat}[H]
\bgroup 
\singlespacing
\begin{tcolorbox}

{\bfseries\large ACI query q4}
\begin{framed}
\begin{lstlisting}[language=SPARQL,basicstyle=\ttfamily\scriptsize]
PREFIX aci: <http://www.aci.it/ontology#>
PREFIX xsd: <http://www.w3.org/2001/XMLSchema#>

SELECT ?veicolo ?stato ?proprieta ?accettazione
?descrizione ?anno ?certificato
WHERE {
  ?veicolo aci:ID_veicolo "<id>"^^xsd:string.
  ?veicolo aci:ha_stato_di_veicolo ?stato.
  ?proprieta aci:proprieta_di ?veicolo.
  ?formalita aci:formalita_presentata_per_veicolo ?veicolo.
  ?formalita aci:data_accettazione_formalita ?accettazione.
  ?formalita aci:anno_formalita ?anno.
  ?formalita aci:descrizione_tipo ?descrizione.
  ?formalita aci:emette_certificato ?certificato.
}
\end{lstlisting}
\end{framed}

{\bfseries\large Description}
\begin{framed}
  Query $q4$ extends $q3$ with the certificate released for the formalities.
\end{framed}
\end{tcolorbox}
\egroup 
\caption{ACI query q4}
\label{lst:aci-query-q4}
\end{lstfloat}


\begin{lstfloat}[H]
\bgroup 
\singlespacing
\begin{tcolorbox}

{\bfseries\large ACI query q5}
\begin{framed}
\begin{lstlisting}[language=SPARQL,basicstyle=\ttfamily\scriptsize]
PREFIX aci: <http://www.aci.it/ontology#>
PREFIX xsd: <http://www.w3.org/2001/XMLSchema#>

SELECT ?veicolo ?stato ?proprieta ?accettazione
?descrizione ?anno ?certificato ?codiceCDP
WHERE {
  ?veicolo aci:ID_veicolo "<id>"^^xsd:string.
  ?veicolo aci:ha_stato_di_veicolo ?stato.
  ?proprieta aci:proprieta_di ?veicolo.
  ?formalita aci:formalita_presentata_per_veicolo ?veicolo.
  ?formalita aci:data_accettazione_formalita ?accettazione.
  ?formalita aci:anno_formalita ?anno.
  ?formalita aci:descrizione_tipo ?descrizione.
  ?formalita aci:emette_certificato ?certificato.
  ?certificato aci:codice_cdp ?codiceCDP.
}
\end{lstlisting}
\end{framed}

{\bfseries\large Description}
\begin{framed}
  Query $q5$ extends $q4$ with the code of the certificate.
\end{framed}
\end{tcolorbox}
\egroup 
\caption{ACI query q5}
\label{lst:aci-query-q5}
\end{lstfloat}


\begin{lstfloat}[H]
\bgroup 
\singlespacing
\begin{tcolorbox}

{\bfseries\large ACI query q6}
\begin{framed}
\begin{lstlisting}[language=SPARQL,basicstyle=\ttfamily\scriptsize]
PREFIX aci: <http://www.aci.it/ontology#>
PREFIX xsd: <http://www.w3.org/2001/XMLSchema#>

SELECT ?veicolo ?stato
WHERE {
  ?veicolo aci:ID_veicolo "<id>"^^xsd:string.
  ?veicolo aci:ha_stato_di_veicolo ?stato.
  ?stato a aci:Stato_rappresentato_valido.
}
\end{lstlisting}
\end{framed}

{\bfseries\large Description}
\begin{framed}
  Query $q6$ asks for vehicles that have a state which is valid.
\end{framed}
\end{tcolorbox}
\egroup 
\caption{ACI query q6}
\label{lst:aci-query-q6}
\end{lstfloat}


\begin{lstfloat}[H]
\bgroup 
\singlespacing
\begin{tcolorbox}

{\bfseries\large ACI query q7}
\begin{framed}
\begin{lstlisting}[language=SPARQL,basicstyle=\ttfamily\scriptsize]
PREFIX aci: <http://www.aci.it/ontology#>
PREFIX xsd: <http://www.w3.org/2001/XMLSchema#>

SELECT ?veicolo ?stato ?numeroTarga ?serieTarga
WHERE {
  ?veicolo aci:ID_veicolo "<id>"^^xsd:string.
  ?veicolo aci:ha_stato_di_veicolo ?stato.
  ?stato a aci:Stato_rappresentato_valido.
  ?stato aci:ha_targa ?targa.
  ?targa aci:numero_targa ?numeroTarga.
  ?targa aci:serie_targa ?serieTarga.
}
\end{lstlisting}
\end{framed}

{\bfseries\large Description}
\begin{framed}
  Query $q7$ extends $q6$ with information about the license plate
  of the vehicles.
\end{framed}
\end{tcolorbox}
\egroup 
\caption{ACI query q7}
\label{lst:aci-query-q7}
\end{lstfloat}


\begin{lstfloat}[H]
\bgroup 
\singlespacing
\begin{tcolorbox}

{\bfseries\large ACI query q8}
\begin{framed}
\begin{lstlisting}[language=SPARQL,basicstyle=\ttfamily\scriptsize]
PREFIX aci: <http://www.aci.it/ontology#>
PREFIX xsd: <http://www.w3.org/2001/XMLSchema#>

SELECT ?veicolo ?stato ?numeroTarga ?serieTarga
?formalita ?codiceFormalita
WHERE {
  ?veicolo aci:ID_veicolo "<id>"^^xsd:string.
  ?veicolo aci:ha_stato_di_veicolo ?stato.
  ?stato a aci:Stato_rappresentato_valido.
  ?stato aci:ha_targa ?targa.
  ?targa aci:numero_targa ?numeroTarga.
  ?targa aci:serie_targa ?serieTarga.
  ?evento aci:determina_stato ?stato.
  ?formalita aci:formalita_genera_evento ?evento.
  ?formalita aci:codice_tipo ?codiceFormalita.
}
\end{lstlisting}
\end{framed}

{\bfseries\large Description}
\begin{framed}
  Query $q8$ extends $q7$ with information about the events
  that determinate the state.
\end{framed}
\end{tcolorbox}
\egroup 
\caption{ACI query q8}
\label{lst:aci-query-q8}
\end{lstfloat}


\begin{lstfloat}[H]
\bgroup 
\singlespacing
\begin{tcolorbox}

{\bfseries\large ACI query q9}
\begin{framed}
\begin{lstlisting}[language=SPARQL,basicstyle=\ttfamily\scriptsize]
PREFIX aci: <http://www.aci.it/ontology#>
PREFIX xsd: <http://www.w3.org/2001/XMLSchema#>

SELECT ?veicolo ?stato ?numeroTarga ?serieTarga
?inizioStato ?fineStato ?formalita
?codiceFormalita
WHERE {
  ?veicolo aci:ID_veicolo "<id>"^^xsd:string.
  ?veicolo aci:ha_stato_di_veicolo ?stato.
  ?stato a aci:Stato_rappresentato_valido.
  ?stato aci:ha_targa ?targa.
  ?targa aci:numero_targa ?numeroTarga.
  ?targa aci:serie_targa ?serieTarga.
  ?evento aci:determina_stato ?stato.
  ?formalita aci:formalita_genera_evento ?evento.
  ?formalita aci:codice_tipo ?codiceFormalita.
  ?stato aci:inizio_stato_del_mondo ?inizioStato.
  ?stato aci:fine_stato_del_mondo ?fineStato.
}
\end{lstlisting}
\end{framed}

{\bfseries\large Description}
\begin{framed}
  Query $q9$ extends $q8$ with more informations about the state.
\end{framed}
\end{tcolorbox}
\egroup 
\caption{ACI query q9}
\label{lst:aci-query-q9}
\end{lstfloat}


\begin{lstfloat}[H]
\bgroup 
\singlespacing
\begin{tcolorbox}

{\bfseries\large ACI query q10}
\begin{framed}
\begin{lstlisting}[language=SPARQL,basicstyle=\ttfamily\scriptsize]
PREFIX aci: <http://www.aci.it/ontology#>
PREFIX xsd: <http://www.w3.org/2001/XMLSchema#>

SELECT ?veicolo ?stato ?numeroTarga ?serieTarga
?formalita ?codiceFormalita ?KW ?cilindrata ?classe
WHERE {
  ?veicolo aci:ID_veicolo "<id>"^^xsd:string.
  ?veicolo aci:ha_stato_di_veicolo ?stato.
  ?stato a aci:Stato_rappresentato_valido.
  ?stato aci:ha_targa ?targa.
  ?targa aci:numero_targa ?numeroTarga.
  ?targa aci:serie_targa ?serieTarga.
  ?evento aci:determina_stato ?stato.
  ?formalita aci:formalita_genera_evento ?evento.
  ?formalita aci:codice_tipo ?codiceFormalita.
  ?stato aci:kw ?KW.
  ?stato aci:cilindrata ?cilindrata.
  ?stato aci:classe_veicolo ?classe.
}
\end{lstlisting}
\end{framed}

{\bfseries\large Description}
\begin{framed}
  Query $q10$ extends $q9$ with more informations about the state.
  This corresponds to one of the competency questions that are
  asked to the system, about technical data of the vehicles.
\end{framed}
\end{tcolorbox}
\egroup 
\caption{ACI query q10}
\label{lst:aci-query-q10}
\end{lstfloat}

\pagebreak

\paragraph{Experimental Setup}
For the purpose of the evaluation we ran the queries over both
the latest version of Mastro $1.0.2$ and Ontop version $3.0$,
using the same specification composed by the ontology and the
mappings exported in R2RML.
The database is managed by an instance of Oracle 12c,
that is accessed remotely over a VPN connection.

The specifications of the platform used for the experiments are the following:
\begin{description}
\item[CPU] Intel Xeon E5-2670 running at 2.60GHz
\item[RAM] 16 GB DDR3 1600 MHz
\item[OS] Ubuntu 17.10 running in a virtualized environment (4 cores)
\end{description}

\begin{figure}[H]
  \centering
  \includegraphics[width=0.9\textwidth]{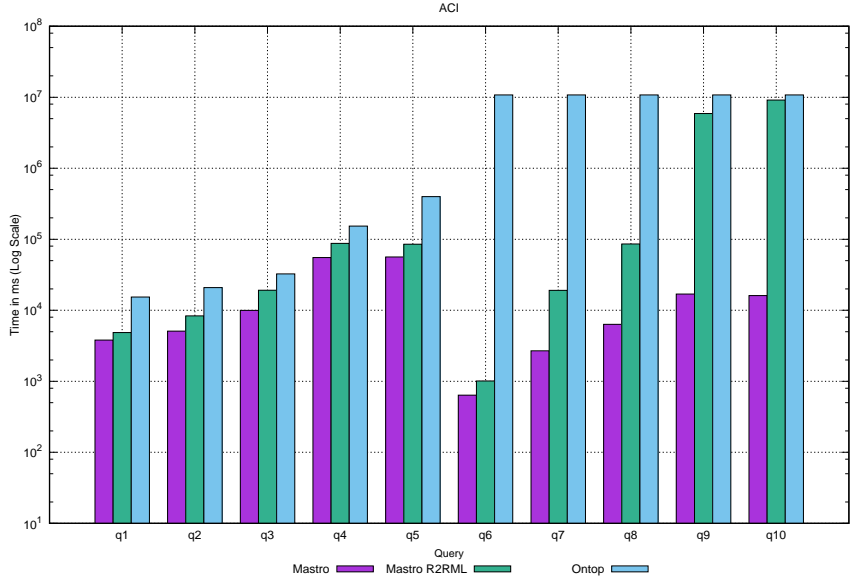}
  \caption{Query Answering over ACI (Execution Time)}
  \label{fig:total-time-aci}
\end{figure}

The experimentation is performed in the following mode:
\begin{itemize}
\item We iterate over the set of queries. At each iteration we pick
  a random query from the set and evaluate it over the system.
  This is done in order to minimize the caching in the DBMS.

  For each execution we store the results and the time it took
  to complete. We consider the combined time needed for evaluating
  the query and processing the set of results.
  Queries are executed sequentially through the
  use of a testing platform designed specifically for this
  task, which accesses directly the internal APIs of the systems.
\item We repeat the process until all queries have been executed
  a fixed amount of time, in this case $5$ executions were performed.
\item Finally, we take the average of the execution times for
  each of the queries in the set.
\end{itemize}
In order to compare the systems under the same setting
we enabled for both reasoning with respect to existential variables.
Our metric of comparison is the total time taken to complete
the execution of the query and to the process the results.
Due to restrictions in the amount of time we were allowed for experiments,
the queries were executed with a timeout of $3$ hours.
As a reference, we report also the time needed by the Mastro system
using the original mapping specification to evaluate the same
set of queries. Figure~\ref{fig:total-time-aci} shows the
overall execution times for both systems. Table~\ref{tab:aci-times}
reports the execution times.

\vfill

\begin{table}[hbtp]
  \caption{Query Answering Times on the ACI Specification (in seconds)}
  \label{tab:aci-times}
  \centering
  \input{aci-times}

\end{table}

\pagebreak

\paragraph{Experimental results}
There are several considerations that can be made by looking at the
results of our experimentation, but the most important one is that
in none of the queries based on the real competency questions
that this solution is designed to answer, Ontop was able
to complete in the time allowed, even though both systems
are using the same specification.
We identified three main reasons for this:
\begin{itemize}
\item The first reason is that, in the
  case of Ontop, the entire rewriting is executed as a single, complex
  SQL query, and the database management system is not able to
  compute an efficient query plan for such large queries.
\item The second reason is that when the optimizations performed
  to reduce the size of the $\mathcal{T}$-mappings fail, because
  there may be missing database constraints, or simply because the
  queries in the mappings contain complex conditions that don't
  allow to apply such optimizations, the system produces rewriting
  containing complex sub-queries, composed of unions of several SPJ queries
  and these types of queries are not evaluated efficiently.
\item The third reason is that, even though representing a very elegant
  solution, having the objects to be constructed directly at the level
  of the data sources, through the use inherently poorly performing
  operations such as string concatenation and type casts, is not
  feasible in real industrial applications.
\end{itemize}
This is due to the fact that the assumption at the base of the Ontop
optimizations, that the axioms in the ontology duplicates the constraints
on the database does not hold in this particular case.
In this case the structure of the data sources reflects the application needs
and not the conceptual reality of the domain of discourse.
When this condition arises, the system is not able to optimize the queries
in the mappings and causes it to produce rewritings that are too difficult to
be dealt with efficiently by the DBMS, where the intermediate views
are composed by complex unions of \emph{select-project-join}
queries.

On the other hand, the approach adopted in Mastro, of splitting the queries
in several, simpler conjunctive queries, even if it can result in an exponential
blowup in the size of the rewriting still enables the system
to complete the task in the time allowed, even if in this case
there are no optimizations performed, as the constraints over the
views are not expressible in R2RML, which cause the time taken
by the system to increase up to almost three orders of magnitude
for the largest query.


%% file: aci-times.tex

\begin{tabular}{c|c|c|c}
  \hline
  \textsc{\bfseries Query} &
  \textsc{\bfseries Mastro} &
  \textsc{\bfseries Mastro R2RML} &
  \textsc{\bfseries Ontop} \\ \hline
  \hline
  $q1$  & $3.808$  & $4.866$    & $15.356$  \\ \hline
  $q2$  & $5.081$  & $8.362$    & $20.874$  \\ \hline
  $q3$  & $9.958$  & $19.139$   & $32.607$  \\ \hline
  $q4$  & $55.310$ & $87.586$   & $153.308$ \\ \hline
  $q5$  & $56.452$ & $85.284$   & $398.358$ \\ \hline
  $q6$  & $0.639$  & $1.011$    & -- (> 3h) \\ \hline
  $q7$  & $2.692$  & $19.113$   & -- (> 3h) \\ \hline
  $q8$  & $6.353$  & $85.915$   & -- (> 3h) \\ \hline
  $q9$  & $16.934$ & $5894.621$ & -- (> 3h) \\ \hline
  $q10$ & $16.104$ & $9123.935$ & -- (> 3h) \\ \hline
\end{tabular}

%% file: sec-conclusions.tex

\section{Conclusions and Future Works}
\label{sec:conclusions}
In this article we have worked on OBDA systems, and in particular,
on the Mastro OBDA system, developed at Sapienza. For such system
we have studied the support of the standard R2RML for realizing
OBDA mappings. We have implemented software components to
support such kind of mappings and our implementation is now
part of the current release of Mastro.
Interestingly, we have seen that we can translated any R2RML mapping into Mastro
mappings, while the contrary is true only if it is acceptable
to lose some efficiency.

Using R2RML mappings we have been able to compare various OBDA systems
against fully standard specifications, where the ontology is specified
in OWL 2, the queries are specified in SPARQL, and
the mappings (to standard SQL DBs) are expressed in R2RML.

In fact, we have seen that only two systems have support fully
the reasoning techniques needed for OBDA solutions: Mastro and Ontop.
Hence, we have concentrated on such two systems.

Our comparison has focused on two benchmarks. The first one is
NPD, a benchmark defined by the Ontop team.
Specifically we have modified
NPD queries focusing on the features that require reasoning support.
The most important result in this case is that it shows how
the idea of moving part of the optimization complexity during an offline stage
can indeed provide an advantage, as it reduces the complexity
of applying optimization algorithms entirely at query execution.

The second benchmark has been based on a full-fledged OBDA solution,
developed within a contract between the department of DIAG at Sapienza,
University of Rome, and ACI Informatica.
In particular, we have used the real sources from ACI as well as the OWL 2
ontology developed by the department for them. With respect to the mappings,
we have translate (losing some efficiency) the real Mastro mappings
into R2RML, so as to allow the comparison with Ontop. Finally, as queries
we used simplified versions of the actual queries used in the
application. The results in this case are quite different.
Most often, Ontop could not handle the queries, while Mastro
could still complete the task in the allowed time slot.
Instead, in the case of Mastro, the approach of separating the rewritings
into single SPJ queries evaluated individually, still enables the
system to complete the task within the allowed time slot, even without
applying any optimization, as these are not expressible in R2RML.

This comparison lead to some interesting results, that can drive
research for future works. In this section we highlight some of them:

\begin{description}
\item[Moving computation to an offline stage:]
  We have seen how moving part of the computation to an offline
  stage can provide large benefits in some cases, not only because
  it could allow the system to produce smaller rewritings by
  applying more extensive optimization techniques that would
  not be possible to execute during query evaluation due to their
  expensive cost in terms of complexity, but also because
  this enables these optimizations to be computed only once,
  without having to recompute them for every execution,
  effectively reducing the time spent during query evaluation.
  This has been already considered in Mastro,
  and a new optimization based on this idea is already under investigation.
\item[Improving R2RML support with database constraints:]
  Database constraints can alone represent a useful source of information
  that can be exploited for optimizations, as can be seen in
  the case of the NPD specification. A possible extension
  to the technique presented in this article for translating R2RML mappings into view
  and ontology mappings can take advantage of these constraints
  to automatically generate data constraints between view mappings.
\item[Combining both approaches:]
  In the case of Ontop, the idea compiling the ontology into the mappings,
  combined with the use of database constraints to simplify
  the resulting mappings, can be indeed a very effective technique
  when the database constraints reflect the knowledge expressed
  by the axioms of the ontology, as they can be combined with techniques that
  avoid the exponential blowup in the size of the produced CQ rewritings
  that can be generated by algorithms such as {\ttfamily PerfectRef},
  and can be computed only once and used for all the unfoldings.
  Although, when such optimization fails to simplify the generated mapping,
  a better approach would be to treat each of the unions as
  a different mapping, and combine it into several, simpler SQL queries
  that can be treated separately, even if it can result in an
  exponential blowup in the number of rewritings, as it is done in Mastro.
  This would at least enable the queries in such case to be efficiently evaluated
  by the database management system.
\end{description}
